\documentclass[garamond,a4paper]{inherent}
\usepackage{xcolor}
\usepackage{graphicx}
\usepackage{subcaption}
\usepackage{booktabs}
\usepackage{array}  
\usepackage{longtable}
\usepackage{xspace}

\usepackage{amssymb}
\usepackage[normalem]{ulem}
\usepackage{enumitem}

\usepackage{hyperref}
\usepackage{url}
\usepackage{relsize}
\usepackage{titletoc}
\titlecontents{lsection}[1.2em]{\sffamily\small}{\contentslabel{1.2em}}{}{\hfill\contentspage}[\vspace{-3pt}]

\usepackage{tcolorbox}
\usepackage{placeins}
\tcbuselibrary{breakable}
\providecolor{InherentCard}{HTML}{F1F5F2}
\newtcolorbox{promptcard}{breakable,colback=InherentCard,colframe=InherentCard,boxrule=0pt,arc=4mm,left=4mm,right=4mm,top=3mm,bottom=3mm}
\usepackage{caption}
\usepackage{listings}
\lstdefinestyle{promptstyle}{%
  basicstyle=\ttfamily\scriptsize,
  breaklines=true,
  breakautoindent=false,
  columns=fullflexible,
  keepspaces=true,
  showstringspaces=false,
  frame=none,
  aboveskip=8pt,
  belowskip=8pt,
  extendedchars=true,
}

\AtBeginDocument{%
  \renewcommand{\sectionautorefname}{Section}%
  \renewcommand{\subsectionautorefname}{Section}%
  \renewcommand{\subsubsectionautorefname}{Section}%
}

\DeclareRobustCommand{\appref}[1]{%
  \begingroup
  \renewcommand{\sectionautorefname}{Appendix}%
  \renewcommand{\subsectionautorefname}{Appendix}%
  \renewcommand{\subsubsectionautorefname}{Appendix}%
  \autoref{#1}%
  \endgroup}

\newcommand{\faradayname}{Faraday\xspace}
\newcommand{\replicaname}{Replica\xspace}
\definecolor{faradaycol}{HTML}{009E73}
\definecolor{faradaycodercol}{HTML}{CC79A7}

\title{Training AI Scientists to Replicate Research}

\newcommand{\authormark}[1]{\textsuperscript{\normalfont#1}}
\newcommand{\eqfirstmark}{\authormark{\relsize{1.3}\textasteriskcentered}}
\newcommand{\infmark}{\authormark{\raisebox{0.15ex}[0pt][0pt]{\S}}}
\newcommand{\eqlastmark}{\authormark{\raisebox{0.15ex}[0pt][0pt]{\relsize{0.5}\textdagger}}}
\newcommand{\authorsep}{{\normalfont\bfseries,}\enspace}

\author{Damon Falck\eqfirstmark\authormark{,1}\authorsep Samer Sabri\eqfirstmark\authormark{,1}\authorsep Anja Surina\authormark{1}\authorsep Thom Foster\authormark{1}\authorsep Anya Sims\authormark{1}\authorsep Sam Devlin\authormark{1}\authorsep \\ Dylan Rogers\authormark{1}\authorsep Tantum Collins\authormark{1}\authorsep Kaloyan Aleksiev\infmark\authormark{,1}\authorsep Louis Kirsch\eqlastmark\authormark{,1}\authorsep Edward Hughes\eqlastmark\authormark{,1}}
\shorttitle{Training AI Scientists to Replicate Research}
\authornote{\eqfirstmark Equal first author. \infmark Infrastructure lead. \eqlastmark Equal last author. \authormark{1}Inherent}
\correspondence{faraday@inherentlaboratories.com}
\date{14\textsuperscript{th} August 2026}

\begin{document}

\maketitle

\begin{abstract}
The replicability of papers is a cornerstone of scientific knowledge, ensuring the reliability of existing results and providing a base for further experiments.
The act of replication typically illuminates details that were previously underspecified, and thus requires similar hypothesis-driven exploration to open-ended research.
In this work, we develop \textit{\replicaname}, a scalable task space for paper replication. 
To provide reward signal, we introduce an auto-generated rubric-based judge that has low noise and agrees with human assessment of replication quality.
We post-train \textit{\faradayname}, a 27B-parameter ``AI Scientist'' agent that leverages coding agents as tools, surpassing the performance of Claude Opus 4.8 and GPT-5.5 on held-out replication tasks.
Qualitative analysis of individual rollouts reveals that \faradayname adopts a more scientifically-principled approach.
We believe that our results provide a stepping stone towards AI agents capable of long-horizon scientific innovation without requiring complex harnesses.
\end{abstract}

\FloatBarrier
\section{Introduction}

\label{sec:introduction}

Science is the search for good explanations about the universe \citep{deutsch2011beginning}. An explanation compresses what we know about reality in a reliable way. 
A good explanation is hard to vary; if new evidence comes to light that contradicts the explanation, that explanation is falsified, rather than easily tweaked to admit the fresh data. Crucial to both reliability and falsification is the idea that scientific experiments ought to replicate: if you run the same experiment, you get the same results, up to the sensitivity of the measuring equipment and uncontrollable stochasticity. Replication, therefore, underpins the edifice of human scientific knowledge.

Remarkably, the sciences face a replication crisis, not least in machine learning \citep{kapoor2022leakage, semmelrock2025reproducibility}.
In principle, LLM-based AI agents offer a scalable resolution to this crisis, especially for research that can be conducted \textit{in silico}.
In practice, however, paper replication poses a challenge to existing AI agents on three fronts.
Firstly, the problem of replicating a paper is underspecified by definition: a paper lossily compresses the research that led to a discovery. 
Secondly, existing AI agents have been heavily trained for well-specified, closed-ended problems~\citep{shao2024deepseekmath, gunjal2025rubrics} whereas replication requires open-ended exploration to infer missing details.
Finally, harnesses like autoresearch \citep{autoresearch} and AlphaEvolve \citep{alphaevolve} do not naturally apply, by virtue of the fact that for a general replication task there is no definite reward on which to hill-climb.
Recent work shows that frontier agents struggle with many scientific aspects of replication, despite their proficiency in engineering \citep{kirgis2026aiagentsconductopenended}. 

In this paper, we train \textit{\faradayname}, an ``AI Scientist'' agent \citep{schmidhuber1991curiosity,muggleton2006artificial,sakanaaiscientist,louiskirschthesis} capable of replicating research papers.
Our LLM-based agent employs Codex GPT-5.5 as a tool, much as human AI researchers use coding agents.
Conceptually, we are training a layer of scientific intelligence that sits above existing coding agents, imbued with an intuition about how to handle underspecified research problems. 
Notably, Faraday is a $27$B-parameter model that directs the work of a model with an estimated $5$T parameters \citep{li2026incompressible} in a way that yields a meaningful performance gain over the larger model alone. 
\newpage

To train \faradayname, we introduce a scalable space of tasks called \textit{\replicaname}.
Each task requires the agent to replicate a figure from a paper with limited time and compute budgets, without seeing the original plot.
Success necessitates a small measure of creativity, in the sense of navigating novel constraints \citep{boden1995creativity,colton2012computational,epstein2026insidethebox}.
A coding agent judge assesses each replication using an auto-generated per-task rubric, validated against expert human rankings, yielding a low-noise reward signal.
\faradayname is produced by post-training Qwen3.6-27B~\citep{qwen2026qwen36} with a turn-level credit variant of GRPO~\citep{shao2024deepseekmath} on the \replicaname train split.

\begin{figure}[t!]
    \centering
    \vspace{-14pt}
    \includegraphics[width=0.95\linewidth]{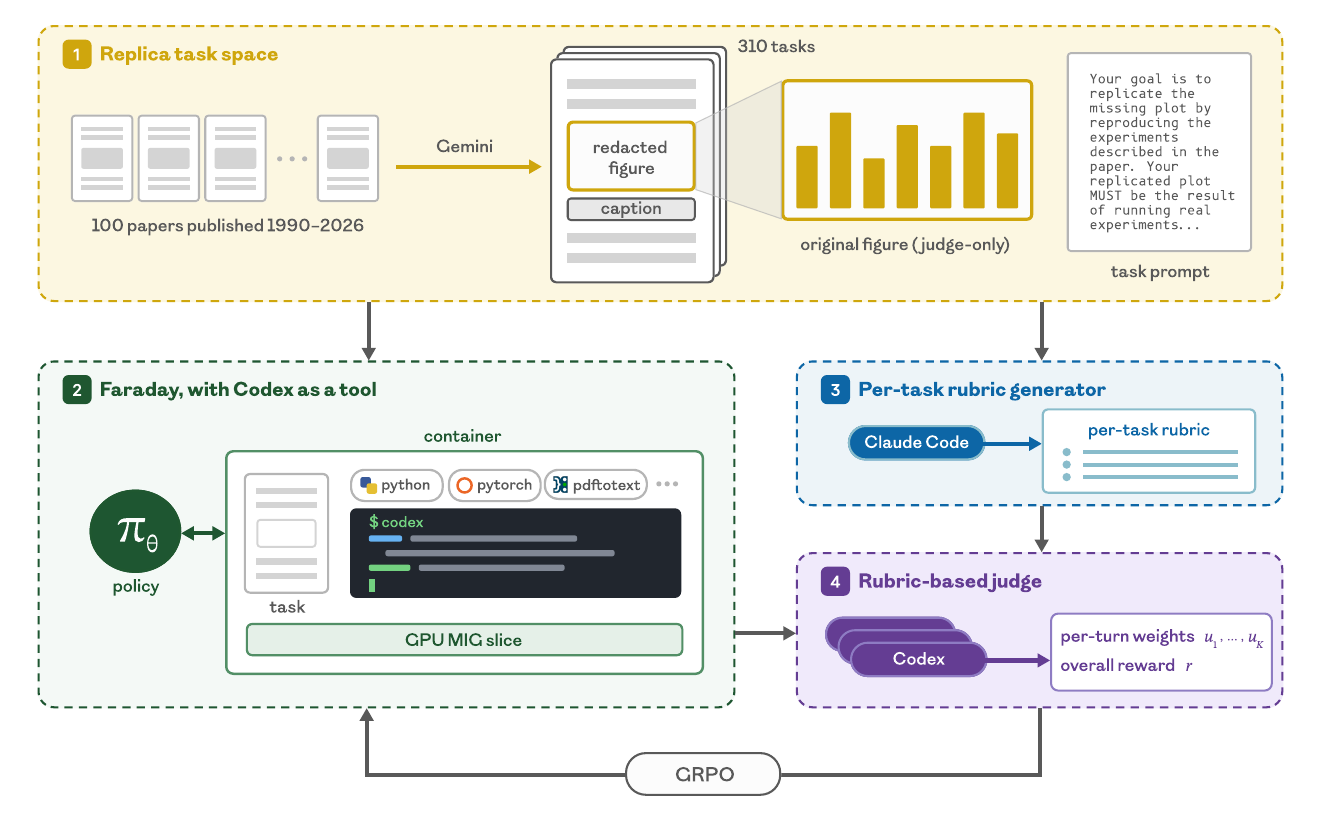}
    \vspace{-8pt}
    \caption{\textbf{We train \faradayname on \replicaname.}
    \textbf{(1)} We construct the Replica task space by curating $100$ ML and AI-for-science papers published between 1990 and 2026.
    For each paper, we generate a set of replication tasks by using Gemini 2.5 Pro to redact individual results figures. Each redacted figure yields one task.
    \textbf{(2)} We generate rollouts on these tasks using our agent Faraday, with access to Codex as a tool for writing code.
    The agent is given a \texttt{containerd} container provisioned with the task prompt, the redacted paper PDF, the Codex binary, various useful research libraries, a one-seventh MIG slice of an H200 GPU, and internet access.
    \textbf{(3)} For each task, we use Claude Opus 4.7 prompted with a meta-rubric to generate a task-specific grading rubric.
    \textbf{(4)} Each rollout is evaluated according to that task's rubric using multiple samples of a Codex-based judge, given access to the rollout's container comprising the generated figure, replication codebase, agent rollout, and ``gold plot'' from the original paper.
    The judge provides an overall reward and per-turn credit assignment weights, which are used to train the Faraday agent using a modified version of GRPO.
    }
    \label{fig:pipeline}
\end{figure}

\begin{figure}[t]
    \centering
    \vspace{-10pt}
    \includegraphics[width=1.0\linewidth]{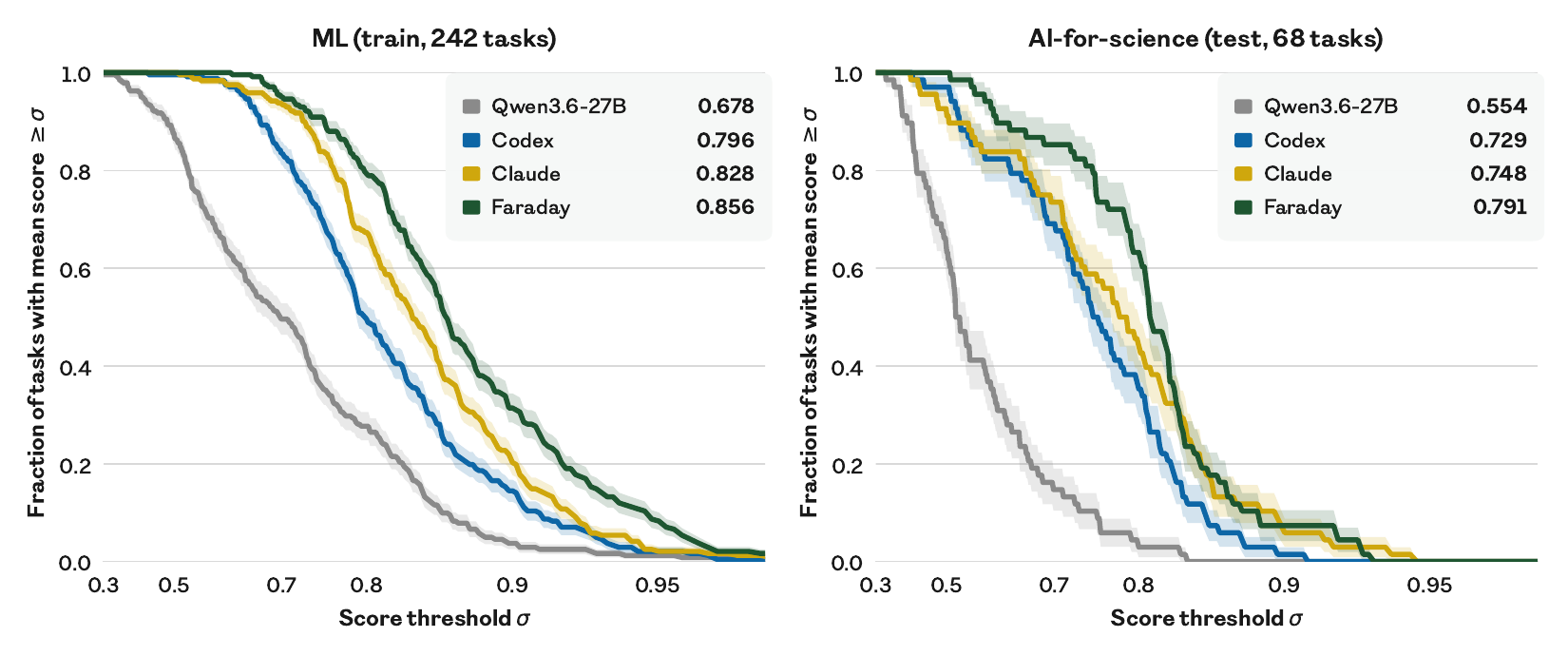}
    \vspace{-10pt}
    \caption{
    \textbf{\faradayname replicates better than frontier coding agents.}
    We plot the fraction of tasks scoring at least $\sigma$, in-distribution (\textbf{left}) and out-of-distribution (\textbf{right}). For each task, we use the mean score over eight evaluation rollouts.
    Bands show $\pm 1$ SEM over tasks.
    \faradayname and Qwen3.6-27B run in the same simple harness, with Codex GPT-5.5 available as a coding tool; they differ only in that \faradayname has been RL post-trained on \replicaname.
    \faradayname's curve lies above baselines at every threshold in distribution, and across almost the whole range out of distribution. Not only is it the strongest agent on average (numbers in legend), it also has a thinner weak tail. 
    As in \citet{airsbench}, the horizontal axis is spaced by a march-of-nines transform, $\sigma \mapsto -\log_{10}(1-\sigma)$.
}
    \label{fig:Faraday_vs_claudex}
    \vspace{-4pt}
\end{figure}

\faradayname outperforms Claude Opus 4.8 (hereafter, Claude) and GPT-5.5 (hereafter, Codex) on $73\%$ of in-\hspace{0pt}distribution ML tasks, and on $60\%$ of held-out AI-for-science tasks, according to our rubric-based judge. On average, \faradayname achieves a $6\%$ improvement over Claude and an $8\%$ improvement over Codex on the test split. Optimising Codex's prompt only marginally diminishes the gap. 
Human experts rate \faradayname as stronger than Claude and Codex on rollouts for which the rubric judge assesses that Faraday has an advantage.
Compared to rollouts from frontier models, \faradayname behaves more like a human scientist: it implements the mechanism behind the claim rather than hard-coding outputs, it scales down in a way that remains faithful to the paper's experimental scope, and it avoids shortcuts that would flatter its own result. In summary, the main contributions of our paper are as follows:

\begin{enumerate}
    \item We introduce \textit{\replicaname}, an automatically generated space of $310$ figure-replication tasks from $100$ machine learning and AI-for-science papers spanning the years 1990--2026 (\autoref{fig:pipeline}).
    \item We provide a recipe for stable GRPO post-training in long-horizon, non-verifiable tasks: a per-task rubric-based judge, multi-sample judge aggregation, and turn-level credit assignment (\autoref{sec:reward} and \autoref{sec:recipe}).
    \item We train \textit{Faraday}, a $27$B-parameter agent that leverages coding agents as tools (CAT), exhibiting greater scientific rigour both quantitatively and qualitatively (\autoref{fig:Faraday_vs_claudex} and \autoref{tab:greatest-hits}).
\end{enumerate}

\section{Related work} 

\paragraph{Rewards for AI Scientists.}
Scientific research is an underspecified problem.
It is context-dependent and admits many different kinds of solutions, and judgement of its quality is subjective.
Automating research is a longstanding goal, often pursued by moving pieces of the research loop inside a learning algorithm, such as the update rule~\citep{schmidhuber1987evolutionary,bengio1992optimization}, the objective~\citep{kirsch2019improving,oh2020discovering}, or the learning algorithm in its entirety~\citep{real2020automl,kirsch2022general}.
These works generally assume a well-defined reward signal against which the meta-learned component can be scored.
Such a signal is hard, if not impossible, to define for the broad goal of scientific research.
Nevertheless, to train AI Scientist agents, it is convenient to compress their behaviour into a scalar-valued reward. 
There are at least two natural ways to produce such a reward: (a) derive a verifiable reward function from existing benchmarks, or (b) judge agent behaviour qualitatively using an LLM. 

The former method is particularly effective when hill-climbing an existing benchmark entails a novel and valuable insight, as was the case for the problems under investigation by FunSearch \citep{funsearch} and AlphaEvolve \citep{alphaevolve}.
The success of such algorithms has spurred much work creating hill-climbable benchmarks for discovery, including in toy settings \citep{discoverybench}, from Kaggle competitions \citep{mle-bench,mledojo}, and based on \textit{in silico} scientific research \citep{mlagentbench,scienceagentbench,mlgym, re-bench,automated-llm-speedrunner,posttrainbench,airsbench}. 

However, this approach suffers a critical limitation: the innovations that agents uncover do not tend to be generalisable.
In other words, they are adaptations but not exaptations \citep{gould1982exaptation}.
There is a ceiling on what can be achieved within this paradigm; indeed, almost no prerequisite to any truly great invention was conceived with that invention in mind \citep{secretan2008picbreeder,stanley2015greatness}. Moreover, considerable human labour was required to construct the aforementioned benchmarks, limiting their scalability for large-scale model training.
\cite{goldie2026procedural} try to resolve both problems by automatically generating a combinatorially huge space from a modest set of hand-designed components, and by explicitly testing generalisation with a meta-train/test split. However, in time, similar problems will emerge at the meta level.

Therefore, we adopt the latter strategy: post-hoc judgement of agent rollouts by an LLM.
A few prior works employ this approach, with the aim of automating research paper generation end-to-end \citep{sakanaaiscientist,cycleresearcher,schmidgall2025agent}.
In these works, the judges mirror peer review, a highly underspecified setting with minimal ground truth and considerable noise.
In contrast, our rubric judge evaluates a more modest, controlled and grounded setting, in which agreement with humans can more easily be established: paper replication.

\paragraph{Replication tasks for AI Scientists.}
Existing replication benchmarks differ in how much of the original work the agent is handed, trading off ease of evaluation with construct validity -- how faithfully they measure replication \citep{construct-validity, construct-validity-in-llms}.
\citet{repro-bench} give the agent a finished reproduction and ask only for a score.
\citet{socsci-repro-bench, core-bench} provide the authors' code and ask the agent to run it and answer questions about the paper.
Various authors supply most of a reference implementation with parts masked out, graded by unit tests \citep{researchcodebench, exp-bench}, code similarity \citep{scireplicate-bench}, or a judge \citep{lmr-bench}.
\citet{from-reproduction-to-replication} sweep this spectrum directly.

We sit at the latter end, where the agent is given the paper and graded by a judge, similar to \cite{paperbench, mlreplicate, prbench, automat}.
Closest are \citet{autoreproduce} and \citet{paper2code}, who also work from the paper against a human-calibrated judge.
We extend this line of work by introducing a larger and more scalable task space, while maintaining the benefits of a per-task rubric \citep{cook2024ticking,goel2025training,gunjal2025rubrics,viswanathan2026checklists,shen2026rethinking,hong2026can}.
Additionally, many of our tasks require the agent to replicate findings under strong resource and time constraints, testing understanding of the method as opposed to blind copying, and necessitating an inventiveness that bridges towards innovative research. 
In concurrent work, \citet{veritas} introduce a complementary task space, extracting a paper's claims and judging each against the evidence from agent-generated experiments across $65$ papers spanning computer science, social science, medicine, and astrophysics.

\paragraph{Training AI Scientists.}
Given a static reward function for discovery, many recent AI Scientist systems have pursued test-time scaling.
These typically rely upon one or more of in-context learning \citep{yang2024large}, evolutionary search \citep{lehman2023evolution, shinkaevolve,aira2}, tree search \citep{aide,aira1,wider-or-deeper}, or test-time training \citep{evotune,cycleresearcher,yang2026reinforcement}.
The test-time improvement algorithm in these works is hard-coded (even if only on the meta level), and thus limited by the biases of the designer \citep{sutton2019bitterlesson}.
Self-modification relaxes this constraint~\citep{schmidhuber1993self,kirsch2022eliminating}, although more recent LLM-based works retain the strictures of fixed language model weights~\citep{zelikman2023self,darwingodelmachine,wang2025huxley,zhang2026hyperagents}.

Other works decompose the scientific method as a hand-designed system of agents with different roles and affordances \citep{ai-researcher, gottweis2025towards, ghafarollahi2025sciagents, ghareeb2026multi}, benefitting from specialisation and division of labour.
However, these modular architectures are somewhat brittle and reductive, and each agent has a restrictive interface, constraining the exploration space in a potentially unhelpful way.
We propose a more flexible setup, in which Faraday is an agent within a \texttt{containerd} container, equipped with a coding agent as a tool (CAT). Our CAT paradigm extends that of \cite{su2025toolorchestra, nielsen2026learning}, in which a smaller model is post-trained to use larger models as tools, to the setting where a tool is a frontier coding agent in its standard CLI harness.

Unlike previous works, we post-train Faraday at scale across a space of $242$ tasks, drawing on the ability of neural networks to generalise, yielding an AI Scientist that can effectively conduct rigorous science out of distribution and without a test-time reward.
This approach draws inspiration from large-scale multi-turn RL without language models \citep{team2021open,ada}, which teaches that training on a vast, smooth, and diverse task distribution produces an agent that can generalise and adapt.
In particular, we succeed at extending GRPO \citep{shao2024deepseekmath} post-training to long-horizon, non-verifiable tasks, a regime known to suffer from instability \citep{xu2025renorm,wang2026arlarena,kim2026horizon}. 
Previous works have used rubric-based judges to generate rewards for multi-turn RL in long-form question-answering \citep{li2026rubricem, shao2025dr} but not for such long-horizon tasks, and not in such a complex environment. 
\cite{xie2026tips} introduce turn-level credit assignment weights from an LLM judge that are similar in spirit but different in formulation to ours, and report negative results. 

\FloatBarrier
\newpage
\section{Methods}
\label{sec:methods}

\subsection{Task space}
\label{sec:replica}

The \textit{\replicaname} task space comprises $242$ training tasks and $68$ test tasks drawn from $100$ well-known ML and AI-for-science papers. Each task requires an agent to replicate one results figure from a paper, given the original paper with the figure redacted, a $60$-minute time limit, and a single one-seventh MIG slice of an H200 GPU. The agent is provided with a \texttt{containerd} container to work in with helpful research libraries pre-installed, access to the internet, a system prompt, and a task prompt (\appref{app:prompts}). Where a paper's experiment cannot be completed within the given time budget, the prompt asks for the most faithful scaled-down version of the underlying experiment. The training tasks are drawn from ML papers from 1990 to 2026. The test tasks are drawn from AI-for-science papers from 2012 to 2026. We choose well-known papers for ease of human rating (\autoref{sec:human-studies}).

Importantly, our tasks are automatically generated, and thus the task space is scalable.
Given a paper, three vision-language stages convert it into a task, powered by Gemini 2.5 Pro~\citep{comanici2025gemini}. 
A scan finds every main-text results plot and its caption, a localisation stage draws its bounding box inside an LLM-verifier repair loop, and the figure is irreversibly redacted from the PDF.
A task is a triple of caption, extracted figure (``gold plot''), and paper with figure redacted. We inspect every task by hand and filter out any that are of low quality, for instance if the figure is insufficiently redacted, if it is not a results plot, or if the caption is incorrectly identified.
Each paper contributes between $1$ and $13$ tasks with median $2$. We address the issue of pre-training contamination in \appref{sec:contamination}.

\subsection{Reward function}
\label{sec:reward}

Paper replication is inherently non-verifiable, especially when scaling down experiments to fit within resource constraints while remaining true to the core claim.
In Replica tasks, the ``gold plot'' (the redacted figure from the original paper) is used to help judge replication attempts, but perfectly reproducing the plot is not the same as a successful replication:
successful replication should also demonstrate strong experimental design, good scientific practice, faithfulness to the original paper, and strategic use of available resources.
Designing a reward signal to train against is therefore a key challenge.
The long-horizon nature of our replication tasks additionally demands that this signal be low-variance across judge samples and consistent across similar rollouts.

\paragraph{Judge rubric generation.} We base our judge on the concept of a rubric, a scoring guide that provides specific criteria for assessing performance on a task. 
Starting from a short, hand-designed meta-prompt, we use Claude Opus 4.7 to auto-generate task-specific rubrics.
We hide the ``gold plot'' from the rubric generator so that the rubric captures the claims of the paper without over-indexing on figure details such as axis ranges, formatting, and exact numerical values.
The judge rubric is also hidden from the model during training, encouraging the model to produce broadly effective replications rather than game the rubric criteria.

The judge rubric covers five dimensions:
(1) how closely the replicated figure visually matches the paper's,
(2) how well the replication supports the paper's scientific claim,
(3) whether the underlying experiment actually implements and tests what the paper describes,
(4) whether the agent makes good use of the compute budget, and
(5) whether the agent acted with scientific integrity, adhering to its instructions and not cheating. 
In our tasks, the time and resource limit means it is often not possible to replicate the figure at full scale.
The rubric generator is instructed to reward agents for producing a faithful scaled-down version.
This is a key feature of the task space: it introduces further underspecification, thereby teaching decision-making skills characteristic of open-ended research. 

\paragraph{Coding agent as a judge.}
We assess rollouts using Codex GPT-5.5 as a judge, prompted with the appropriate per-task rubric.
The judge is given access to the same workspace and compute resources as the agent, including the redacted paper, all the tools the agent had, the replication codebase, and git history (containing the final plot generated), and the full interaction trace of the rollout, as well as the ground-truth ``gold plot'' from the original paper.
The judge is given $10$ minutes to explore these materials and form a judgement on each dimension of the rubric.
Each criterion receives a continuous score between $0$ and $1$, and these per-dimension scores are averaged to give an overall score for the rollout.
Crucially, this approach allows the judge to examine fully and potentially re-execute the agent's code to understand its process and check the robustness of its claims.
During training, we sample from the judge three times for each rollout to reduce variance, and additionally instruct the judge to generate credit assignment weights for each agent turn; see \autoref{sec:recipe}.

\subsection{Human studies}\label{sec:human-studies}

We collect human rankings of agent rollouts to assess the extent to which our rubric-based judge captures human research taste.
For each rollout, the human expert is given access to the relevant paper with the figure redacted, the ``gold plot'' and its caption, a transcript of the rollout, and the git repository generated by the agent, including agent instructions, code and outputs, and the plot(s) that resulted from the agent's experiments. Humans are asked to rank either three or six rollouts from best to worst.
Participants are provided with simple instructions for how to rank rollouts; see \appref{app:human-prompt}.
Importantly, they are instructed to follow their own best judgement about what they would expect a faithful replication to look like, so as to capture human tacit knowledge.

Participants are asked to justify their ranking, so as to encourage a principled and consistent approach \citep{mcdonnell2016rationales}. We also ask participants to explain what effect the ``gold plot'' shows, and to suggest a correct methodology to reproduce it. Given the complexity of the task, we select participants from a pool of ongoing or completed PhDs from top research universities, preferring participants who have published at least one paper at the main conference track of ICML, ICLR, or NeurIPS. We pay each participant £150 per task, with bonuses of £125 paid upon the completion of the fifth and eighth tasks. In total we collect 117 rankings from 20 participants. \appref{app:human-studies} describes how we choose the tasks and rollouts for participants to rank.

\subsection{Simple harness}
\label{sec:harness}

An agent harness provides an interface between an LLM (tokens-in/tokens-out) and an environment (action-in/state-out). In our setting, the environment is an interface to a container. We design an agent harness for Faraday based on three principles: it should be simple and interpretable; it should be maximally permissive, allowing the agent the same context and affordances a human would have when undertaking AI research; and its tool set should be minimal, deliberate, and legible to contemporary models.
Faraday's scientific capabilities are improved by changing its policy weights, not by complexifying its harness. We report the harness system prompt in \appref{app:system-prompt}.   

\paragraph{Affordances and context.}
The agent acts via five function-calling tools: \texttt{apply\_patch}, \texttt{read\_file}, \texttt{list\_dir}, \texttt{grep\_files}, and \texttt{shell}. Faraday can detach background processes using the shell tool, allowing it to take many turns while running several commands in parallel. 
The tool interface is a subset of the Codex CLI schema \citep{openai2025codex}, reimplemented in Python.
The conversation is a linear, append-only history with no compaction.
A turn's tool calls execute concurrently and their results are appended in call order.
Context overflow, exceeding the per-turn $16$K token limits, and inference errors end the rollout, and the partial rollout is judged like any other.
A rollout otherwise ends when Faraday replies without tool calls or when its wall-clock time is exhausted.

\paragraph{Coding agent as a tool (CAT).}
Faraday is provided with a frontier coding agent to use as a tool.
A wrapper script runs the Codex CLI non-interactively. 
Faraday can invoke this script through its \texttt{shell} tool, receiving a rendered transcript of the coding agent's commands, outputs, and messages with per-step timings.
Successive invocations resume the coding agent's previous session by default.
However, Faraday can choose to reset context or run multiple coding agents in parallel.
The wrapper script enforces a deadline, configurable by Faraday per-request. If the deadline is exceeded, a partial transcript is returned. 
The coding agent model is a runtime parameter; we use GPT-5.4 mini for most of the training, and GPT-5.5 in the final stage and for evaluation.

\FloatBarrier
\subsection{Post-training recipe}
\label{sec:recipe}

\begin{figure}[t]
\vspace{-4pt}
    \centering
    \begin{subfigure}[c]{0.48\linewidth}
        \centering
        \includegraphics[width=\linewidth]{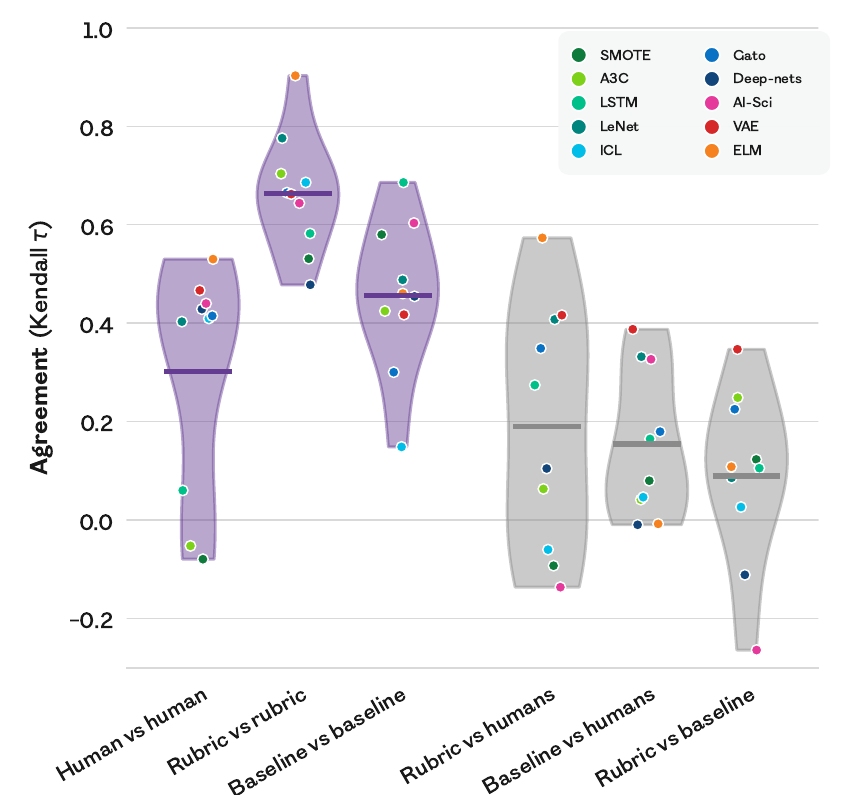}
        \label{fig:rubric-judge-a}
    \end{subfigure}
    \hfill
    \begin{subfigure}[c]{0.48\linewidth}
        \centering
        \includegraphics[width=\linewidth]{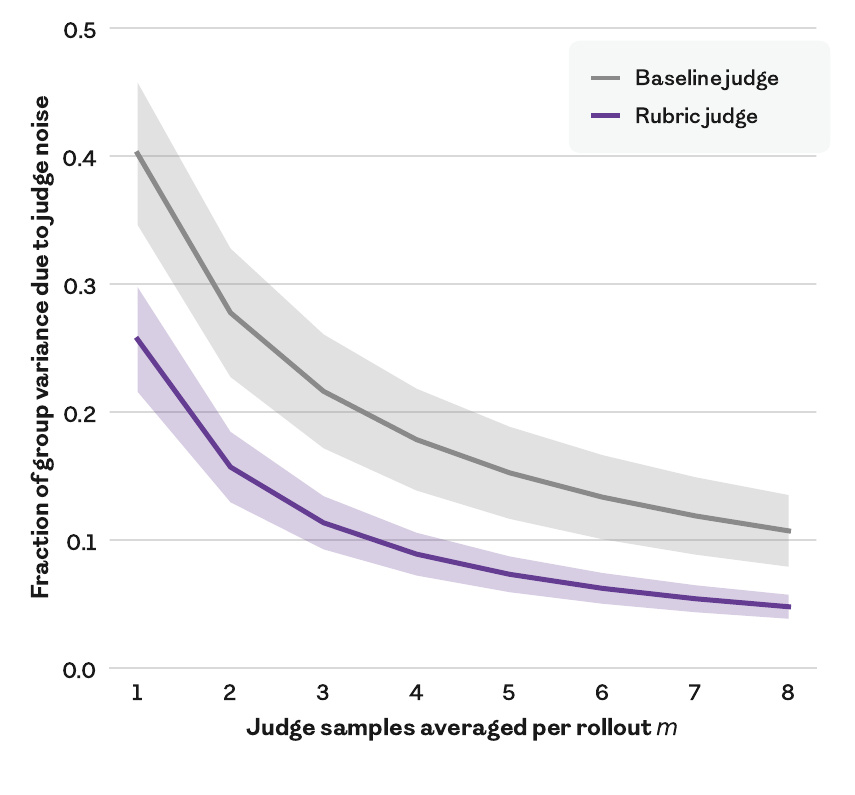}
        \label{fig:rubric-judge-b}
    \end{subfigure}
    \caption{
    \textbf{Our rubric judge achieves higher human agreement and lower noise than the baseline judge.}
    We compare our per-task rubric-based prompt against a simpler baseline prompt which does not vary across tasks.
    Codex GPT-5.5 uses the prompt to judge rollouts sampled from Claude, Codex, and Faraday.
    \textbf{(left)} We select tasks whose rollouts maximise disagreement between the rubric judge and the baseline judge. We ask expert humans to rank the same rollouts based on their intuition for what constitutes a good replication. We measure agreement using Kendall $\tau$, a rank correlation between two orderings (+1 = identical, −1 = reversed). Two independent draws of the rubric judge agree more closely ($0.66$) than two draws of the baseline judge ($0.46$) or two humans ($0.30$). The rubric judge agrees more closely with humans ($0.19$) than the baseline judge ($0.15$). Dots are individual tasks, listed in \autoref{tab:judge-comparison}.
    \textbf{(right)} We sample $16$ GRPO groups of 8 rollouts each uniformly across training steps $430$--$461$, scoring every rollout eight times with each judge.
    For each group, we calculate the fraction of the within-group score variance that is caused by judge noise, as a function of the number $m$ of judge samples averaged per rollout. Bands show $\pm 1$ SEM over the 16 groups. The rubric judge is less noisy at every $m$. In particular, eight baseline judge samples are required to reduce the noise share to the level obtained with three rubric judge samples.
    }
    \vspace{-6pt}
    \label{fig:rubric-judge}
\end{figure}

To obtain \faradayname, we post-train Qwen3.6-27B~\citep{qwen2026qwen36} in the Faraday harness on the \replicaname task space, using a modified version of GRPO~\citep{shao2024deepseekmath}.
We use LoRA fine-tuning~\citep{hu2022lora} with rank $128$ and $\alpha=128$; we train adapters on all linear projections, with a $128$K-token context window and a constant learning rate of $6 \times 10^{-6}$. We find this context length to be sufficient for the \replicaname tasks given the 1-hour time limit used during training.
Each Adam \citep{kingma2014adam} optimiser step draws a batch of $10$ tasks, with eight rollouts apiece, from the $242$-task \replicaname train split. Tasks are sampled so that every batch spans the corpus' year range evenly, and each epoch visits every task exactly once, so no single era of science dominates any update. \appref{app:post-training} provides further training details and \appref{app:infra} describes our infrastructure.

\paragraph{Long-horizon stability.}
Our post-training requires long-horizon RL in a non-verifiable domain, a setting known to be prone to instability and collapse.
Two sources of this instability are the high variance of the reward signal and uniform credit assignment.
To address this, we make two train-time modifications to our judge.
First, we compute the rollout-level reward using the mean of three independent judge evaluations.
Second, we instruct the judge to produce turn-level weights attributing credit over the rollout's turns.
The judge produces a weight distribution $u_k$ over turns $k$ which is then normalised such that $\sum_k u_k n_k = \sum_k n_k$, where $n_k$ is the number of tokens in turn $k$, such that we do not change the overall reward scale. The normalised weights are averaged turn-wise over the three judge draws, which preserves the normalisation. The weight for the corresponding turn is then used to scale the per-token advantage during GRPO. In this way, credit is redistributed within a rollout without changing the overall magnitude of the update. \appref{fig:turn-attribution} explores empirically how our method assigns credit within each rollout. These techniques helped to achieve stable training (see ablations in \appref{app:ablations}).

\FloatBarrier
\section{Results}

\begin{figure}[t]
\centering
\begin{subfigure}{0.48\linewidth}
\centering
\includegraphics[width=\linewidth]{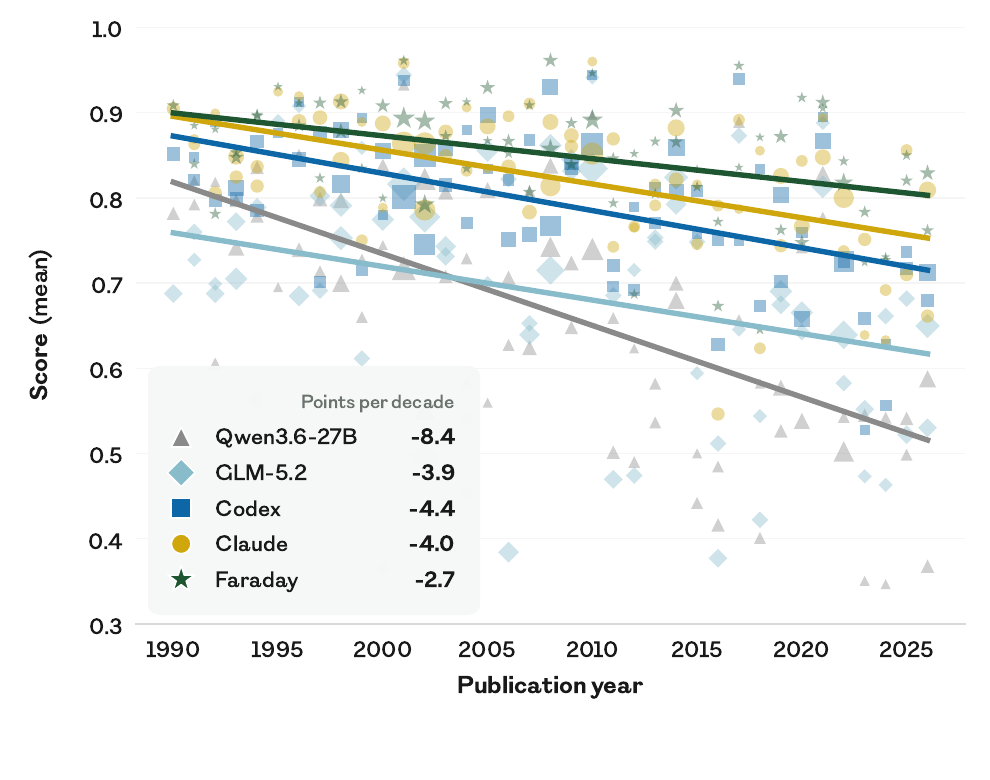}
\label{fig:baselines-a}
\end{subfigure}
\hfill
\begin{subfigure}{0.48\linewidth}
\centering
\includegraphics[width=\linewidth]{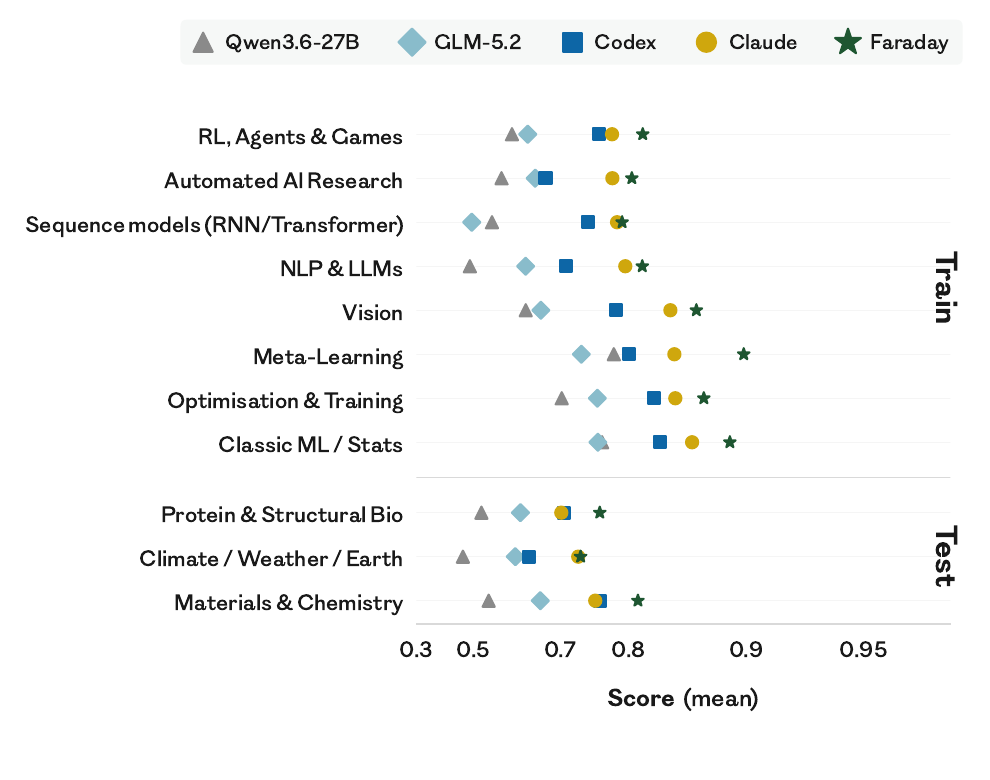}
\label{fig:baselines-b}
\end{subfigure}
\caption{
\textbf{Frontier coding agents do not saturate \replicaname.}
\textbf{(left)} Papers written more recently are harder to replicate.
Each point represents the mean rubric score across tasks for one paper from the Replica train split, sized by the number of tasks that paper yields.
Lines are least-squares fits.
\textbf{(right)} Difficulty varies across research topics, and \faradayname leads on every one.
The top block shows the train split (ML tasks), and the bottom block shows the test split (AI-for-science tasks).
Each point represents the mean rubric score over tasks in the given research topic.
}
\label{fig:baselines}
\end{figure}

\subsection{Rubric judge reliably captures human taste} \label{sec:results_human_study}

We conduct a human study to establish alignment between our rubric judge and human taste. We report full methodology and results in \appref{app:judge-comparison}. We introduce a baseline judge with the same coding agent model (Codex GPT-5.5) but a constant prompt across tasks that mirrors the prompt given to human participants. 
Human rankings correlate better with our per-task rubric judge than with a baseline judge (\autoref{fig:rubric-judge}, left). However, there are several tasks on which our rubric judge disagrees with humans, suggesting room for improvement in future work. Our rubric judge ranks more consistently than both the baseline judge and humans. 
Furthermore, the fraction of group variance that arises from judge noise rather than between-rollout signal is lower for the rubric judge (\autoref{fig:rubric-judge}, right). It is therefore a better candidate for use as a reward in GRPO.

\subsection{\replicaname tasks are challenging for frontier agents}
\label{sec:baselines}
We run frontier coding agents on Replica and find that they do not saturate the task space. For Claude Opus 4.8 and GPT-5.5 baselines, we run the model in the Claude Code and Codex harnesses respectively, with extra-high thinking effort. For Faraday, we pin the thinking effort of its Codex tool to extra-high, to ensure a fair comparison. For the GLM-5.2 baseline, we run the model with max thinking effort in the Claude Code harness, as it is the best reported harness for TerminalBench \citep{zai2026glm52}. Every agent receives the same task materials and the same $60$-minute single-GPU budget, and is scored by the same rubric judge. We run eight rollouts per task per agent.
Within a task, rollout scores are reduced to a single per-task score by taking the mean.

Claude Opus 4.8 is our strongest baseline. Task performance decreases with publication year for every agent. We speculate that more recent papers are harder to replicate because there is less density of information about them in the pre-training dataset, and because they tend to use higher compute resources, and so determining an appropriate and successful scale-down is more challenging. Task difficulty also varies between research topics, and the per-topic rankings are consistent among baseline agents, with NLP and LLM papers hardest and classical machine learning and statistics easiest. AI-for-science papers are generally harder to replicate than ML research papers across agents, possibly because they require integrating experimental expertise from different domains. Faraday's base model and harness before RL is the weakest of all, and is the fastest to degrade with recency. 

\subsection{\faradayname replicates better than Claude and Codex}\label{sec:results_quantitative}

We compare \faradayname against baselines across the entire Replica task distribution (\autoref{fig:Faraday_vs_claudex}). We achieve a comprehensive uplift in performance compared to the base Qwen model, on both train and test tasks. In distribution, \faradayname outperforms both Claude and Codex on $73\%$ of tasks. Out of distribution, \faradayname outperforms Claude and Codex on $60\%$ of tasks. Since the held-out papers span research areas \faradayname never trained on, the behaviour it acquired is not memorisation of a specialised procedure but a transferable way of approaching the underspecified task of paper replication. On both train and test, Faraday's advantage is an upward shift of the whole distribution. \autoref{fig:baselines} (right) shows that the gap is consistent across different subdomains. Decomposing the judge score into its sub-dimensions reveals that \faradayname is stronger than baselines when it comes to experimental depth, claim reproduction, and visual fidelity, and matches Claude on scientific integrity and implementation fidelity (\autoref{fig:perf-profile-by-rubric}).

\begin{figure}[t]
\vspace{-6pt}
    \centering
    \begin{subfigure}{0.48\linewidth}
    \centering
    \includegraphics[width=\linewidth]{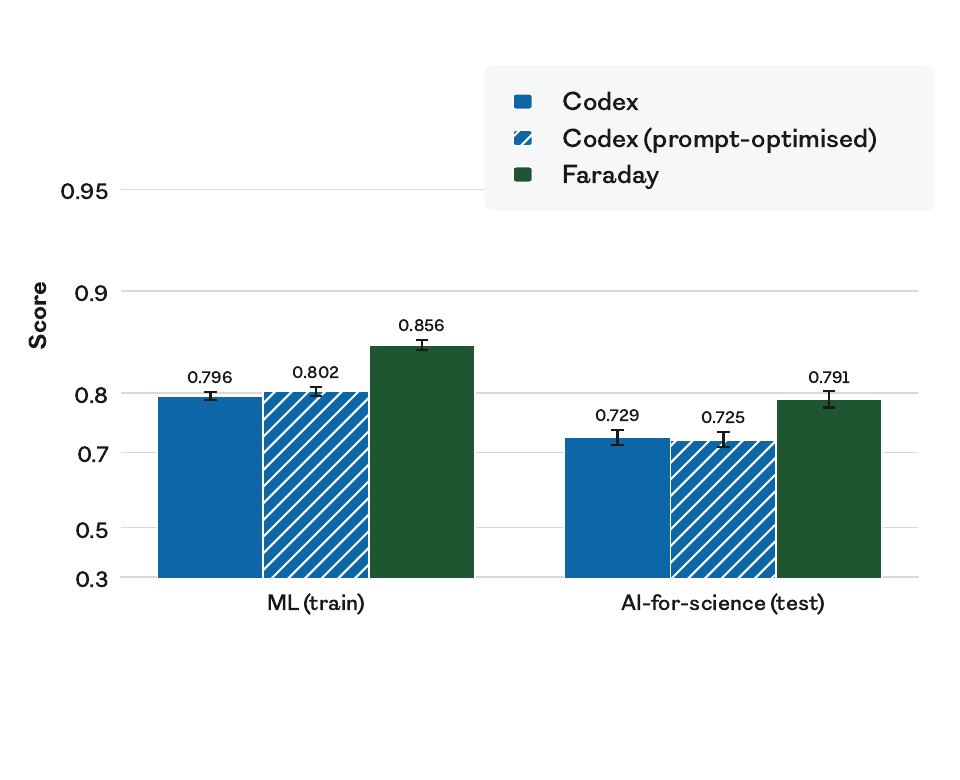}
    \end{subfigure}
    \begin{subfigure}{0.48\linewidth}
    \centering
    \includegraphics[width=\linewidth]{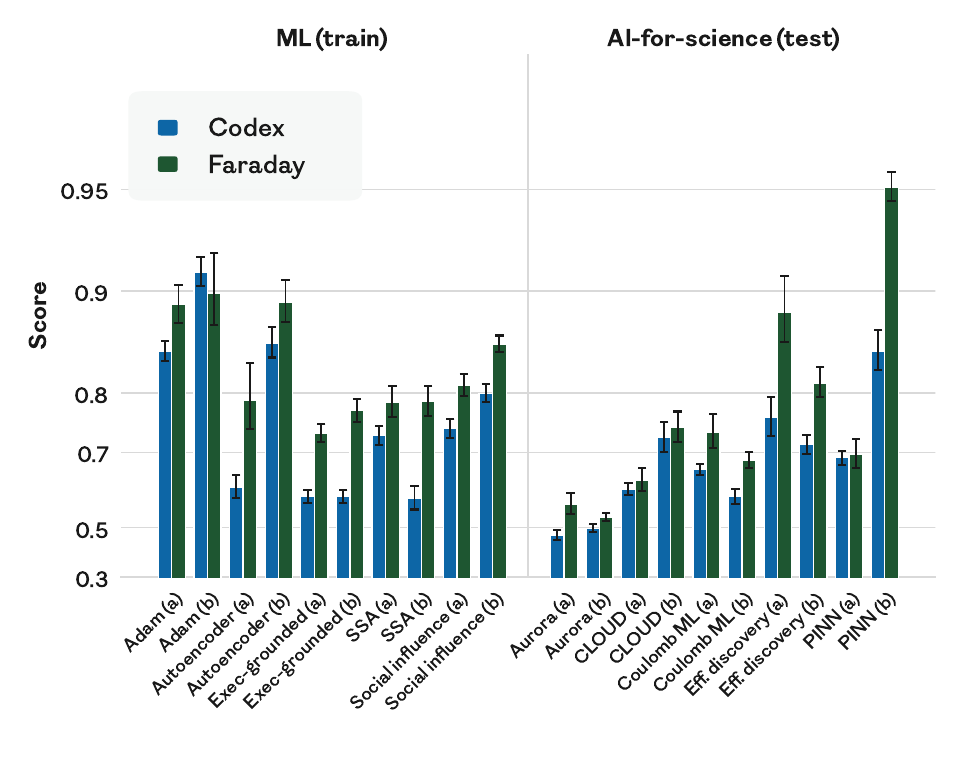}
    \end{subfigure}
    \caption{
    \textbf{Prompt-optimised Codex remains weaker than Faraday, and Faraday generalises to tasks that require innovation.}
    \textbf{(left)} The mean rubric score over all tasks (and over eight rollouts per task) for Faraday is higher than for Codex with both the default baseline prompt, and a prompt automatically optimised in-context for one epoch over the train split.
    \textbf{(right)} On twenty counterfactual variants of tasks from ten papers, five from the train split and five from the test split, Faraday leads on mean rubric score over eight rollouts in almost all cases. For each task, one variant swaps the dataset and one changes the claim.
    In both plots, bars show $\pm 1$ SEM. 
    }
    \label{fig:promptopt}
\end{figure}

\newpage
To test whether \faradayname's advantage can be obtained by prompting alone, we run $24$ generations of automated prompt optimisation on the Codex baseline.
Similar to \faradayname's training, each generation samples $10$ training tasks with eight rollouts per task.
Claude Opus 4.8 then rewrites the prompt based on all previous rollouts in the filesystem, including the judge's feedback.
We compare the final prompt against Codex and Faraday in \autoref{fig:promptopt} (left).
The optimised prompt does not perform meaningfully better than the original prompt, thus the gap to Faraday is retained.
The optimised prompt (\appref{app:optimised-prompt}) identifies the specific failure modes seen in the rollouts, but without success:
the gain from post-training does not appear to be reachable by prompting. 

\subsection{Faraday is qualitatively a more rigorous scientist}

\label{sec:results_qualitative}

To understand how \faradayname improves on the Claude and Codex baselines, we examine by hand the individual rollouts for which \faradayname's rubric judge score exceeds the best Claude and Codex score by the largest margins. 
\autoref{tab:greatest-hits} showcases representative examples.
Two patterns recur. First, 
\faradayname implements the mechanism an experiment is designed to test, whereas the baseline hardcodes the expected output or falls back on an oversimplified method, failing to replicate the main claim of the figure.
Second, \faradayname is more thorough in the scope of its experiments, reproducing more of the original experiment without unnecessary omissions.

Furthermore, we examine the discovery process that occurs within Faraday rollouts.
We sample nine tasks uniformly at random from the Replica test split, and for each task select the strongest of eight Faraday rollouts.
We identify the moments of insight when the best score until that point is exceeded and use Claude Opus 5 to label the insights. 
We see a similar accumulation of knowledge as in AI Scientist systems built with evolutionary harnesses (\autoref{fig:rollouts-faraday}). 
However, unlike previous systems, Faraday has no special hand-coded harness, does not change its harness at test time, and does not have access to the rubric judge reward. 
In other words, Faraday has learned to value insights intrinsically. 

Finally, we run a human study to assess to what extent humans prefer \faradayname over Claude and Codex, specifically focussing on rollouts in which the rubric judge deems that \faradayname holds a strong advantage. We report full methodology and results in \appref{app:agent-comparison}. Of the $41$ rollouts examined, humans prefer \faradayname over both Claude and Codex in $29$, significantly more than chance. This result suggests that the rubric judge accurately but not perfectly captures the characteristics of good replication, at least when it comes to the best-performing samples from \faradayname. Importantly, the design of our study does not allow us to draw any conclusions as to whether humans prefer \faradayname over Claude and Codex on average. Gathering conclusive evidence for this preference would require a larger-scale study across a randomly selected set of tasks, and is an important direction for future work. 

\begin{table}[p!]
    \centering
    \caption{\textbf{\faradayname behaves more like a rigorous scientist.} Via human inspection, we qualitatively analyse tasks with the largest margin between \faradayname and the best performing run among Claude and Codex, grouped by whether the paper lies inside \faradayname's training distribution (ML) or outside it (AI-for-science). 
    }
    \label{tab:greatest-hits}
    \small
    \renewcommand{\arraystretch}{1.15}
    \begin{tabular}{@{}p{0.24\linewidth} p{0.70\linewidth}@{}}
        \toprule
        \textbf{Task} & \textbf{Difference in approaches} \\
        \midrule
        \midrule
        \multicolumn{2}{@{}l}{\textit{\textbf{ML (train)}}} \\
        \midrule
        \textbf{Darwin-G\"odel Machine}\newline\cite{darwingodelmachine}, Fig.~4.\newline
        &Figure~4 tests whether DGM-discovered improvements transfer across models, benchmarks, and programming languages. \faradayname implements the paper's evolutionary self-improvement procedure, building an archive of mutated agents and transferring its best agent. The baseline instead hard-codes a putatively discovered agent, bypassing the search that the experiment is meant to demonstrate.\\
        \addlinespace[3pt]
        \textbf{Learning Precise Timing with LSTM Recurrent Networks}\newline\cite{gers2002learning}, Fig.~12.
        & Figure~12 shows trained peephole LSTMs generating periodic rectangular functions. At the first try, model training in both rollouts does not converge. \faradayname's coding agent tries to patch the issue by hand-crafting the network in place of training it, which \faradayname stops and constructs a workable training recipe instead. The Codex baseline instead steers the network towards the desired behaviour through its initialisation and auxiliary losses, and then selects a favourable checkpoint that shows the desired result.
        \\
        \addlinespace[3pt]
        \textbf{Voyager}\newline\cite{wang2024voyager}, Fig.~8.
        & Figure~8 tracks intermediate progress on unseen crafting tasks, testing whether skills acquired during exploration transfer zero-shot. \faradayname runs a dedicated skill-acquisition phase that learns the skills and transfers them to held-out tasks, replicating the mechanism the figure measures. The best Claude rollout instead supplies a hard-coded, pre-populated library including target-solving skills, so the central library-transfer mechanism is specified by hand and not tested.
        \\
        \addlinespace[3pt]
        \textbf{The AI Scientist}\newline\cite{sakanaaiscientist}, Fig.~2.
        & Figure~2 tests GPT-4o paper-reviewing ablations. \faradayname ran the paper-reviewing pipeline at a meaningful scale, producing reviews for several times more papers than the Codex baseline. It also followed the method more closely: five self-reflection rounds and an area-chair meta-review, versus one critique prompt and an average over five reviews. Accordingly, \faradayname's reflection loop improves the accuracy while the Codex one barely moves it. \\
        \addlinespace[3pt]
        \midrule
        \midrule
        \multicolumn{2}{@{}l}{\textit{\textbf{AI-for-science (test)}}} \\
        \midrule
        \textbf{ChemVAE}\newline\cite{chemvae}, Fig.~4.
        & Figure~4 shows that Gaussian-process search in a learned molecular latent space finds higher-scoring molecules and visualises optimisation paths. \faradayname more faithfully implements the paper's method: a generative model that turns any point in its learned space back into a molecule, so optimised points become molecules the model writes, with a fallback to the nearest known molecule only when a decode is invalid. Codex simplifies some parts of the model, and due to this, its representation cannot be turned back into a molecule, so every
        molecule in its figure is retrieved from the dataset rather than generated by the model. \\
        \addlinespace[3pt]
        \textbf{GNoME} \newline \cite{gnome}, Fig.~3.
        & Figure~3 shows that more pretraining data improves models that predict atomic forces for unseen materials. \faradayname repeats each scaling-law training-set size five times and reports the spread; the best Codex rollout uses one seed per scaling point with no reported uncertainty. Only \faradayname implements the figure's robustness test of fine-tuning at low temperature and evaluating at high temperature; Codex draws both sets from one generator call that takes no temperature argument, so the claimed shift reported on the axis is not supported. \\
        \bottomrule
    \end{tabular}
\end{table}

\FloatBarrier
\section{Discussion}

\paragraph{Towards innovation.}
On the face of it, replicating a figure from a paper is not an especially creative endeavour.
Most obviously, replication produces a figure that looks quite like the original, assuming that the method replicates.
But if one examines the process, rather than the output, replication becomes a stepping stone towards innovation.
The skills that allow Faraday to fill in vaguely-specified details may be the very same skills that would allow it to advance the state of the art by designing its own experiments \citep{deutsch2011beginning, muthukrishna2016innovation, heyes2018cognitive, bhoopchand2023learning}.
Indeed, many human researchers start their careers by learning to replicate existing results and, upon mastering that, are better placed to conduct original research.
Replication is the first step in a curriculum of increasing underspecification towards innovation. 

Inspired by such considerations, we assess how well Faraday generalises to ``imagined'' replications (\autoref{fig:promptopt}, right).
We ask Claude Opus 4.8 to generate two variants of five randomly selected papers from each of the Replica train and test splits:
(a) making the same claim as the original figure but using a different dataset or environment and
(b) making a different claim from the original figure in the same setting (\appref{app:innovation-tasks}).
We evaluate Faraday and Codex GPT-5.5 on these tasks, and score them with our rubric judge, noting that the task interface itself has not changed.
Faraday's rollouts are preferred to Codex's by the judge on $19$ of the $20$ tasks.
In a weak sense, Faraday not only replicates better than a frontier model; it also innovates better.
However, we must caution that our rubric judge was never validated on imagined tasks, and so future work is warranted to validate this claim. 

\paragraph{Coding agent as a tool (CAT).}
It is perhaps surprising that we succeed in training such a small model to better direct the activities of a model at least two orders of magnitude larger.
Moreover, training the outer agent need not be prohibitively expensive in inference tokens for the inner tool.
After training with a weaker coding agent as a tool, one can substitute a more powerful coding agent at evaluation time and achieve an uplift in performance (\autoref{fig:coder-swap}).
The skills \faradayname acquires -- deciding what to investigate, scoping experiments to a budget, and judging a replication -- compound with advances in frontier coding models.
One might hope that a single post-trained outer agent can track the frontier as better models are released, at least over some time period.
Establishing the optimal cost-benefit tradeoff between the sizes of the inner and outer agents is an interesting topic for further study. 

The success of the CAT paradigm has implications for both capabilities and safety.
In the realm of AI Scientist agents, we offer an approachable alternative to harness construction.
History teaches us that encoding capabilities in the weights of a neural network, rather than expressing them in code, is more flexible and generalisable in the long run.
With an eye to safety, our results demonstrate successful oversight of a more powerful model by a less powerful one \citep{amodei2016concrete, bowman2022measuring, kenton2024scalable}.
Moreover, the reasoning traces of the open-weights model can be inspected, unlike those behind the closed-weights API surface.
Note that nothing in the CAT paradigm requires the outer agent to remain the smaller model; it is an empirical question whether the demands of scientific judgement must match or exceed those of engineering execution in the long run. 

\paragraph{Beyond verifiable rewards.} To capture the scientific abilities that underpin open-ended research, we necessarily move away from well-specified tasks with verifiable rewards. A side effect of this reorientation may be that agents are less exposed to incentives for reward hacking during post-training \citep{baker2025monitoring}. Defining a verifiable reward necessitates specifying an evaluation procedure in foresight, which becomes a fixed target for manipulation. By contrast, judging entire rollouts in hindsight is a moving target. Indeed, we observe Faraday acting with greater scientific rigour and faithfulness than frontier agents, completing tasks as intended rather than reproducing figures performatively. It remains to be seen whether training on open-ended tasks can scalably ameliorate reward hacking.

\paragraph{Generalisation.}
In training Faraday, we deliberately limit the scope of the tasks to short time horizons and limited GPU resources.
Our motivation is twofold: first, pragmatism in achieving sufficient throughput for RL per unit wall-clock time; second, a belief that the ability to experiment quickly and efficiently with minimal versions of research ideas is a valuable transferable skill.
It is natural to wonder whether Faraday can generalise to larger resources, similar to those used for experiments in the original papers.
To assess this, we select one figure from each of eight papers whose replication we estimated to require fewer than eight hours and eight B300 GPUs.
We provide appropriate resources to Faraday and to Claude Opus 4.8 and evaluate them on these scaled-up tasks (\appref{app:scale-generalization}).
We find that Faraday exceeds the performance of Claude on average and in five out of the eight tasks, suggesting generalisation. 
Clearer validation of the rubric judge on larger-scale tasks, together with a larger number of such tasks, would be required to make a stronger claim. We further discuss scaling in the supplementary discussion (\appref{app:supplementary-discussion}). 

\paragraph{Community engagement.}
Paper replication is a public good, strengthening the scientific foundations upon which future insights can be built.
Progress on replication is particularly timely, since the paper review system is beginning to strain under the weight of AI-assisted research \citep{gartenberg2026more}.
Indeed, high-quality paper replication tools may well help to ground AI-assisted reviewing in the future.
If you have ideas for how you might use Faraday in your work, we would be delighted to hear from you at \href{mailto:faraday@inherentlaboratories.com}{\texttt{faraday@inherentlaboratories.com}}.

As an early step towards real-world validation of Faraday's usefulness, we obtained feedback from the authors of four papers in the Replica task space \citep{rupp2012fast, chemvae, reed2022generalist, sakanaaiscientist} on Faraday's replication of one figure from their paper.
On the one hand, the authors were impressed by parts of the replication (``part b and c look very good'', ``the reflexion implementation looks correct''), by the inventiveness of the agent (``nice and clever toy task design'') and by fidelity to the original work (``the agent's implementation more closely follows equation (1) in the paper''). 
On the other hand, some simplifications did not make sense (``the problem selected is probably too easy''), parts of the write-up were poor (``paragraph on pre-training dataset design is particularly bad'') and code slop is off-putting (``calculations contain unnecessarily convoluted code'').

\paragraph{Conclusion.} In summary, we have created an intelligence layer with a modicum of research taste, sufficient to extend the capabilities of frontier agents.
This is, however, the tip of the iceberg when it comes to imbuing agents with the ability to enrich scientific research as peer collaborators with humans.
Stepping from replication towards innovation sharpens the problem of underspecification, and deepens the need to develop systems with good judgement.
And cultivating creative human-machine teams will require AI taste in code, experiments, theories, collaboration, and organisational design. 

\section*{Ethics statement}

\paragraph{Awareness of limitations.}
Paper replication by AI agents holds great promise, but remains at an early stage. While Faraday successfully replicated claims from a number of papers, it failed in several cases where we have confidence that the original result was obtained rigorously and reported honestly. We do not claim that any of Faraday's failures to replicate published work suggest fundamental problems with the original research. Even as and when AI systems demonstrate sufficiently strong performance to serve as reliable judges of replicability, it will remain important for humans to cultivate the skills necessary for paper replication and to inspect the results of agents such as Faraday.

\paragraph{Human studies.}
We assessed whether our procedures for human data collection would require review by an external board and concluded that this was not necessary since the information elicited consisted only of professional judgements and did not feature sensitive personal data, and participation would pose no risk. We worked with a combination of people within our pre-existing professional networks and experts sourced by third-party data providers. We made clear to participants the purpose of the project. Participants were compensated irrespective of whether their ratings cleared our internal filtering. 

\paragraph{Conflict of interest.}
Some of the papers included in our corpus were written by authors of this paper and by individuals we know personally. Some work in the corpus comes from institutions whose commercial models we used for the research presented here. We applied the same automated pipeline to all items in the corpus and in no way altered scoring treatment for our own prior work or that of our acquaintances.

\paragraph{Technical safety.}
Not all applications of scientific insight benefit society. Prior work on the use of AI to automate or accelerate scientific research notes accurately that these capabilities may empower malicious human actors and/or increase the dangers associated with misaligned AI systems. For this paper, we selected \textit{in silico} tasks that we judge unlikely to cause harm, and we constrained Faraday in terms of both time and compute. Faraday did not have access to any physical lab equipment, although it did have internet access. 

\section*{Acknowledgements}
We thank Shi Dong, Alex Goldie, Matt Henderson, Akarsh Kumar, Chris Lu, Clare Lyle, and Jimmy Secretan for valuable comments on an early version of this manuscript.
We thank Sergio Gomez, Jos\'{e} Miguel Hern\'{a}ndez-Lobato, Chris Lu, and Matthias Rupp for providing feedback on the quality of replication of their papers.

\bibliography{iclr2026_conference}
\bibliographystyle{iclr2026_conference}

\FloatBarrier
\newpage

\appendix
\numberwithin{figure}{section}
\numberwithin{table}{section}
\renewcommand{\thefigure}{\thesection.\arabic{figure}}
\setcounter{figure}{0}

\renewcommand{\thetable}{\thesection.\arabic{table}}
\setcounter{table}{0}

\section*{\fontsize{15.5}{18}\selectfont Appendix}

\FloatBarrier
\section{Generalisation}
\label{app:generalization}

\subsection{Full-scale replication}
\label{app:scale-generalization}

\faradayname is trained to complete scaled-down replications of a single figure given a one-hour time limit and a one-seventh MIG GPU slice. More precisely, during early stages of training, \faradayname is given a time limit of $30$ minutes, which is increased to one hour for the later stages. Previous work has indicated that a horizon curriculum may induce effective generalisation to longer horizons than experienced during training \citep{kim2026horizon}. Here, we test Faraday's ability to generalise to completing full-scale replications, given the time and compute necessary to do so.

We select eight replication tasks from outside \faradayname's training distribution, filtered using Claude Opus 4.8 such that at most eight hours and eight B300 GPUs should be sufficient to replicate the figure. Of the selected tasks, five are from AI-for-science papers \citep{xie2017cgcnn, chithrananda2020chemberta, ramsundar2015massively, xu2025cloud, bhattacharjee2026srcgcnn} and three are from ML papers \citep{gu2026asymgrpo, lin2026selfregress, yuan2026diversity}. 
Three of the eight papers were first made publicly available after the knowledge cutoff for Claude Opus 4.8~\citep{anthropic2026models}, GPT-5.5~\citep{openai2026gpt55} and Qwen3.6-27B~\citep{qwen2026qwen36}.
For this evaluation, we also increase Faraday's context limit to its maximum $256$K, up from $128$K during training; we choose the eight-hour time limit because this is the approximate time horizon allowed by the increased context limit without compaction.

\begin{figure}[h!]
    \centering
    \includegraphics[width=0.6\linewidth]{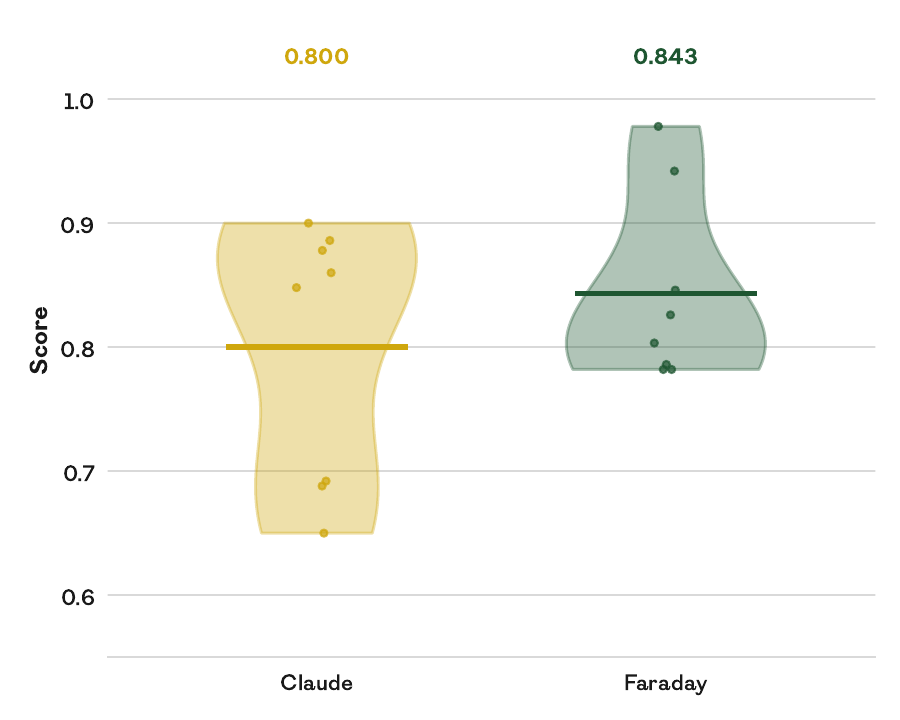}
    \caption{
    \textbf{Faraday outperforms Claude on full-scale replications.}
    On eight held-out replication tasks, Claude and Faraday are evaluated on their ability to replicate the experimental results at full scale, with access to up to eight hours and eight B300 GPUs (as estimated to be necessary to complete a full replication without any scale-down). Faraday performs better than Claude according to our rubric judge (horizontal rules represent the means over the tasks), and outperforms it on five of eight tasks.
    }
    \label{fig:ai4sci-scale-gen}
\end{figure}

We use Claude Opus 4.8 to estimate how many hours and how many B300 GPUs (up to the cap of eight hours and eight GPUs) should be necessary to fully replicate one figure from each paper without any scale-down. We then run one rollout for each task using Faraday and Claude, under the time and compute resources estimated. We find that Faraday outperforms Claude on average according to our rubric judge (\autoref{fig:ai4sci-scale-gen}). In other words, Faraday generalises to longer time horizons and larger compute resources. The caveat is that we did not validate our rubric judges with human ratings on rollouts at this scale, an important step for future work.

\FloatBarrier
\newpage

\subsection{Stronger coding agent tool}
\label{app:coder-swap}

\begin{figure}[h!]
    \centering
    \includegraphics[width=0.62\linewidth]{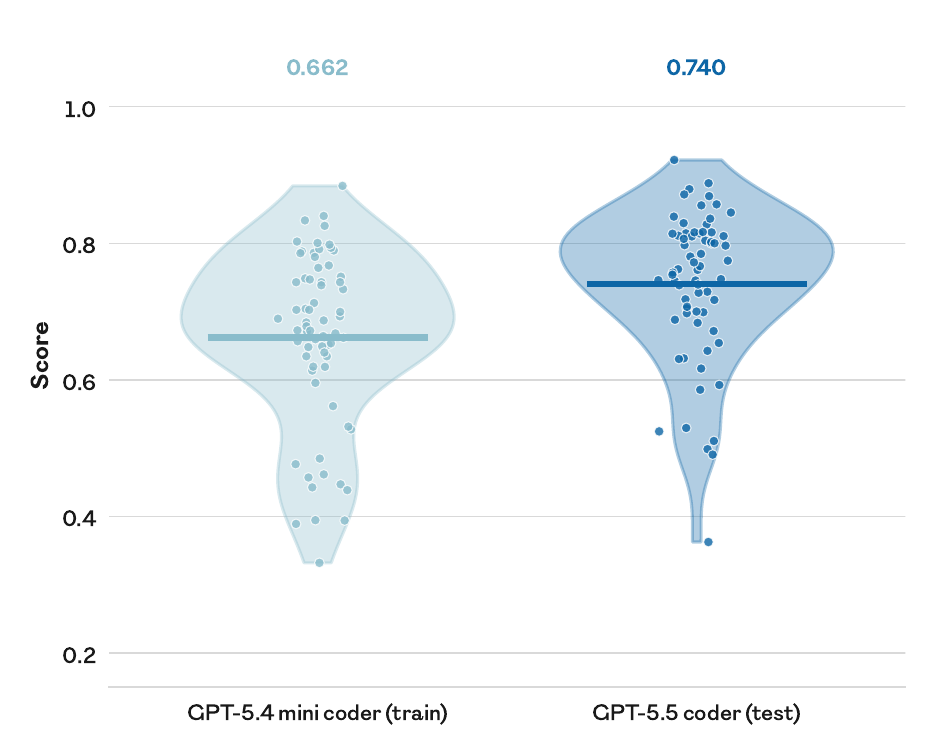}
    \caption{
    \textbf{The trained AI Scientist is not specialised to its train-time coding agent.}
    The last checkpoint from the Faraday training lineage that is trained entirely with GPT-5.4 mini as the coding agent performs better on held-out AI-for-science tasks when the coding agent is swapped out for GPT-5.5, demonstrating that Faraday can generalise to use a stronger coding agent without the need for retraining.
    Each point is an individual task (mean of four rollouts), and the horizontal rules are the means across tasks.
    }
    \label{fig:coder-swap}
\end{figure}

Since the capabilities of frontier coding agents increase frequently, it would be useful if \faradayname were able to make effective use of stronger coding agent tools than it was trained on. To evaluate this generalisation, recall that \faradayname is trained initially with GPT-5.4 mini as the model backing the Codex tool. We take the last checkpoint from \faradayname's lineage which was trained only with GPT-5.4 mini as a tool, and we evaluate it on the Replica test split using first GPT-5.4 mini and then GPT-5.5 as the coding agent model. \autoref{fig:coder-swap} demonstrates that this partially trained version of \faradayname makes effective use of the stronger coding agent to boost its scores on the tasks.

\FloatBarrier
\newpage
\section{Ablations}
\label{app:ablations}

\subsection{Turn-level credit assignment}
\label{app:turn-credit}

\begin{figure}[h!]
    \centering
    \includegraphics[width=0.9\linewidth]{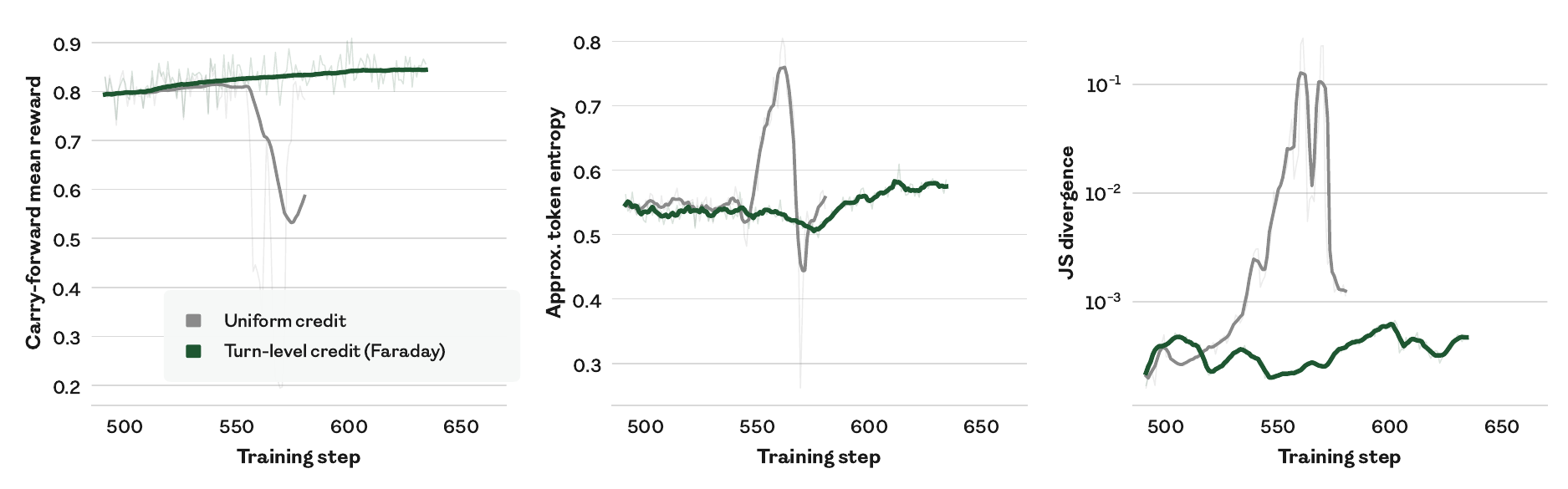}
    \caption{
    \textbf{Turn-level credit assignment stabilises training.}
    Starting from a late checkpoint in Faraday's training lineage, the removal of our turn-level credit assignment results in the rapid destabilisation and collapse of training.
    \textbf{(left)} With turn-level credit assignment, the carry-forward mean reward (the mean over all tasks of the most recent reward achieved in that task) rises steadily, whereas with uniform credit assignment it collapses after $50$ steps. \textbf{(centre)} Around the same time, the token entropy of the policy trained without turn-level credit assignment spikes and then collapses. \textbf{(right)} Leading up to the collapse, the Jensen--Shannon divergence between the generation policy and the training policy (which differ due to asynchronous training) begins to increase, eventually growing by two orders of magnitude. We find this to be a common precursor to such collapses.
    }
    \label{fig:turnjudge-stability}
\end{figure}

\subsection{Coding agent as a tool}
\label{app:codex-tool}

\begin{figure}[h!]
    \centering
    \includegraphics[width=0.75\linewidth]{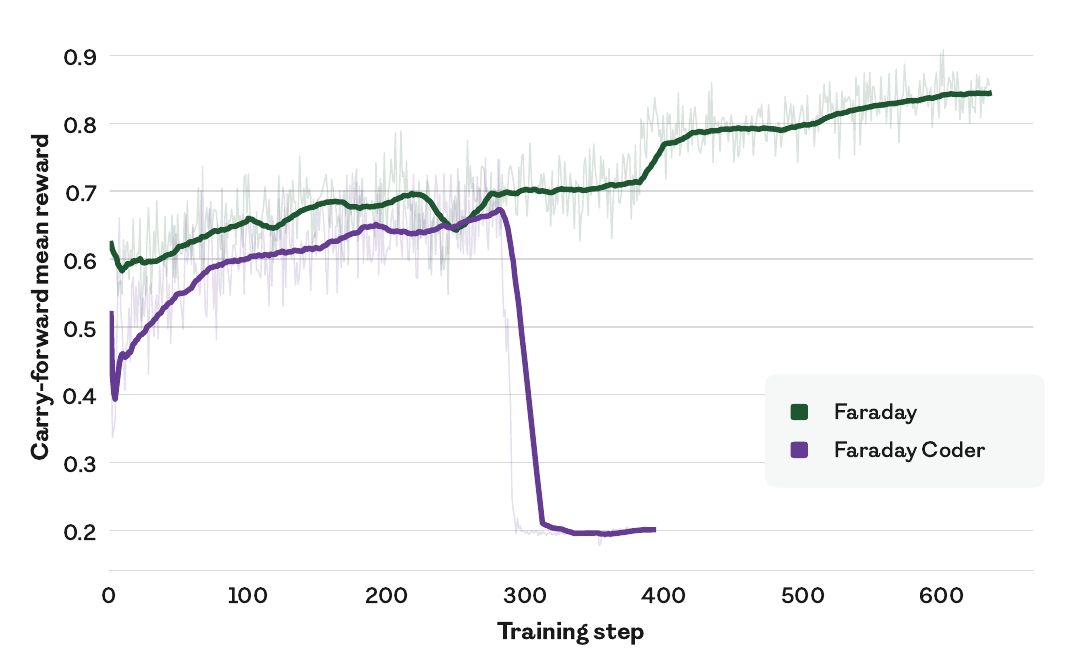}
    \caption{\textbf{The coding agent tool is important for best performance.} We train the Qwen3.6-27B model from scratch using the same hyperparameters as the final tailpatch of the Faraday training lineage but without access to the coding agent tool (``Faraday Coder''). Training collapses after approximately $300$ steps, after performing more weakly at equal step count to the Faraday lineage. Notably, during the pre-collapse period, Faraday Coder had twice the time horizon ($60$ minutes) of Faraday ($30$ minutes), and still performed consistently worse. This suggests that the ceiling for the coder model is lower than that for the researcher model. As in \autoref{fig:turnjudge-stability} (left), the curves show the carry-forward mean reward, with the ghosted curves representing the per-step mean reward.}
    \label{fig:base-model}
\end{figure}

\FloatBarrier
\newpage
\section{Analyses}
\label{app:additional-results}

\subsection{Credit assignment distribution}
\label{app:credit-dist}

\begin{figure}[h!]
    \centering
    \includegraphics[width=0.9\linewidth]{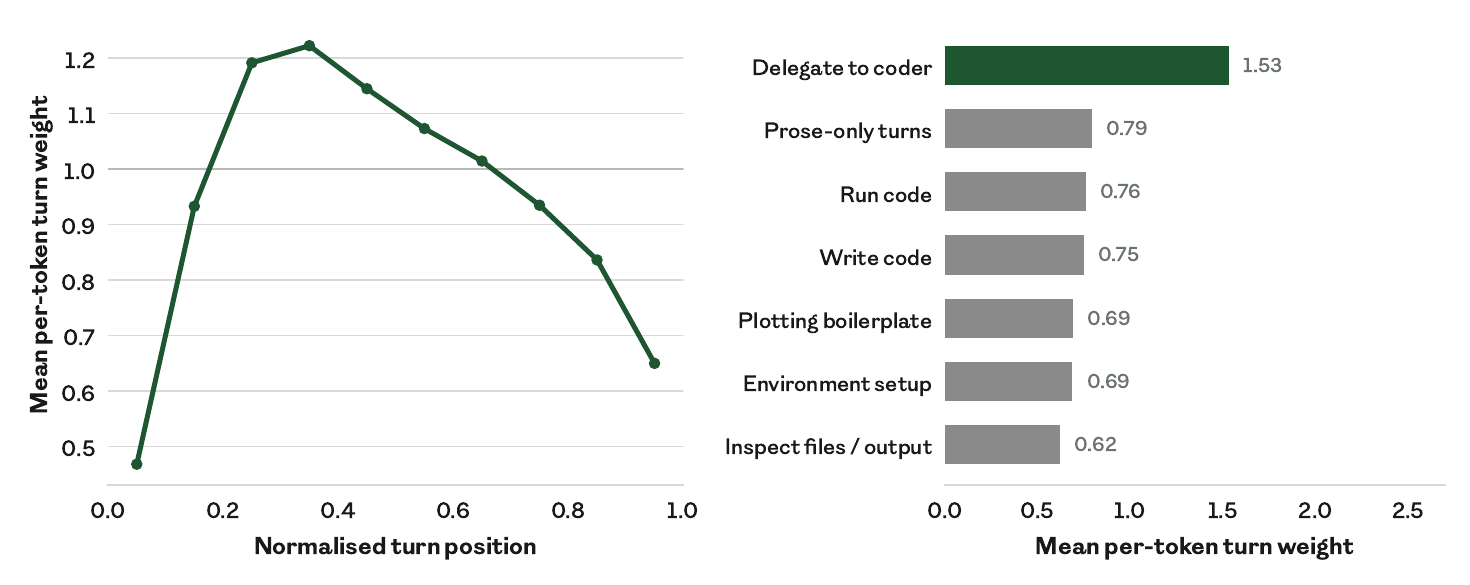}
    \caption{
    \textbf{The judge spreads credit non-uniformly across each rollout.}
    We draw data from $577{,}585$ turns of the Faraday training lineage with turn-level credit assignment enabled (every turn from steps $491$--$635$). 
    \textbf{(left)} There is a concentration of credit in the early to middle stages of a rollout, where load-bearing decisions are most commonly made.
    \textbf{(right)} More weight is given to turns that delegate to the coding agent tool, capturing the importance of appropriate delegation. Turn types are assigned post-hoc by a regular-expression match on the turn text.
    }
    \label{fig:turn-attribution}
\end{figure}

\newpage

\subsection{Scores by rubric dimension}
\label{app:perf-by-rubric}

\begin{figure}[h!]
    \centering
    \includegraphics[width=0.9\linewidth]{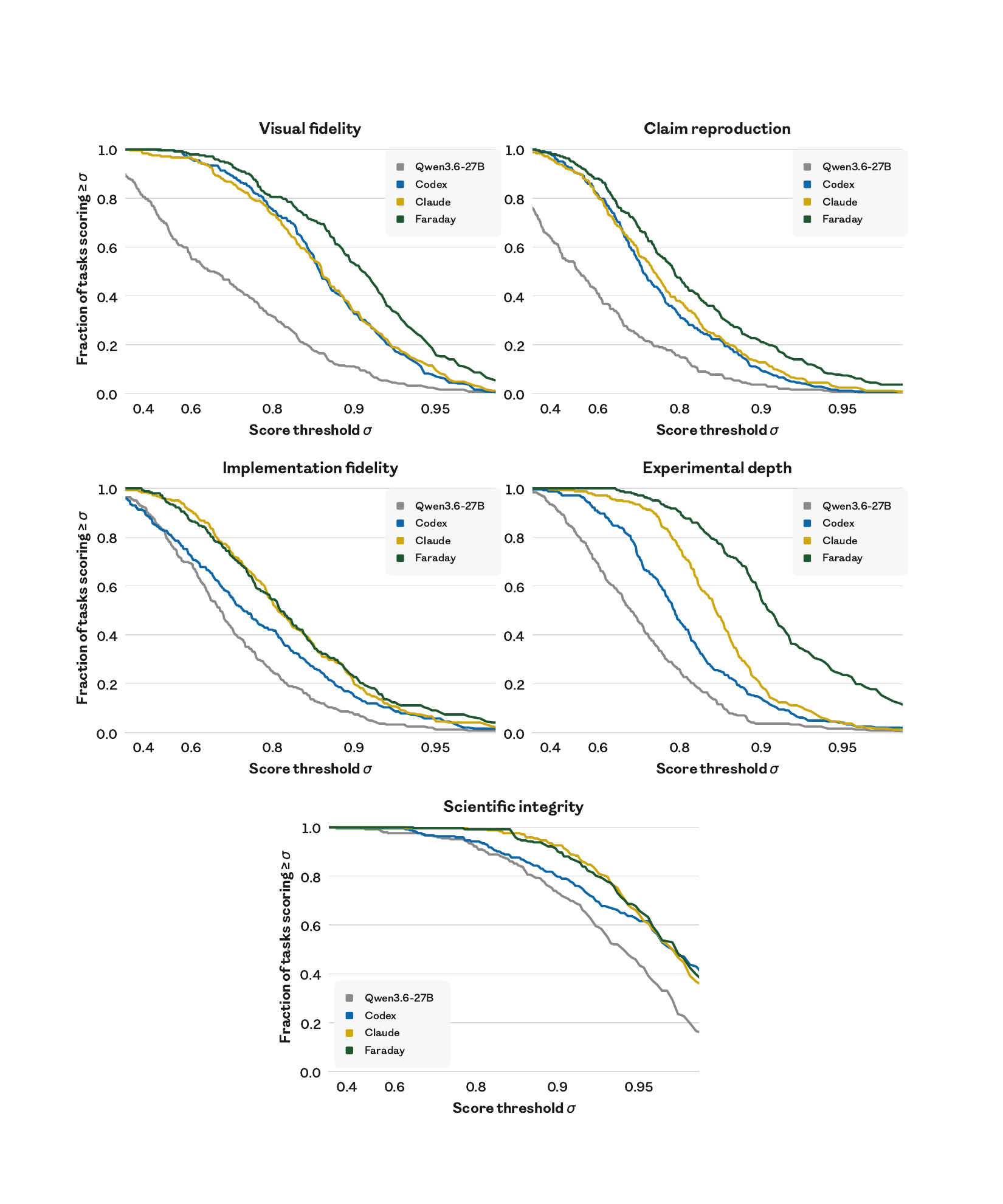}
    \captionsetup{width=\textwidth}
    \caption{
    \textbf{\faradayname's advantage is concentrated in experimental depth and claim reproduction.}
    The left-hand panel of \autoref{fig:Faraday_vs_claudex} is split out into the five score dimensions of our rubric judge. As in that figure, each panel shows the fraction of the $242$ tasks in the Replica train split with a mean score over eight rollouts of at least $\sigma$ in the corresponding rubric dimension. We omit the SEM for visual clarity.
    Faraday's replications consistently have more experimental depth, better claim reproduction, and higher visual fidelity to the original figure. Faraday approximately matches Claude in implementation fidelity (faithfulness to the paper's methodology) and scientific integrity (not cheating while completing the task). See \autoref{sec:reward} for a description of the rubric dimensions.
    }
    \label{fig:perf-profile-by-rubric}
\end{figure}

\FloatBarrier

\newpage
\subsection{Within-rollout behaviour}

\begin{figure}[h!]
    \centering
    \includegraphics[width=0.85\linewidth]{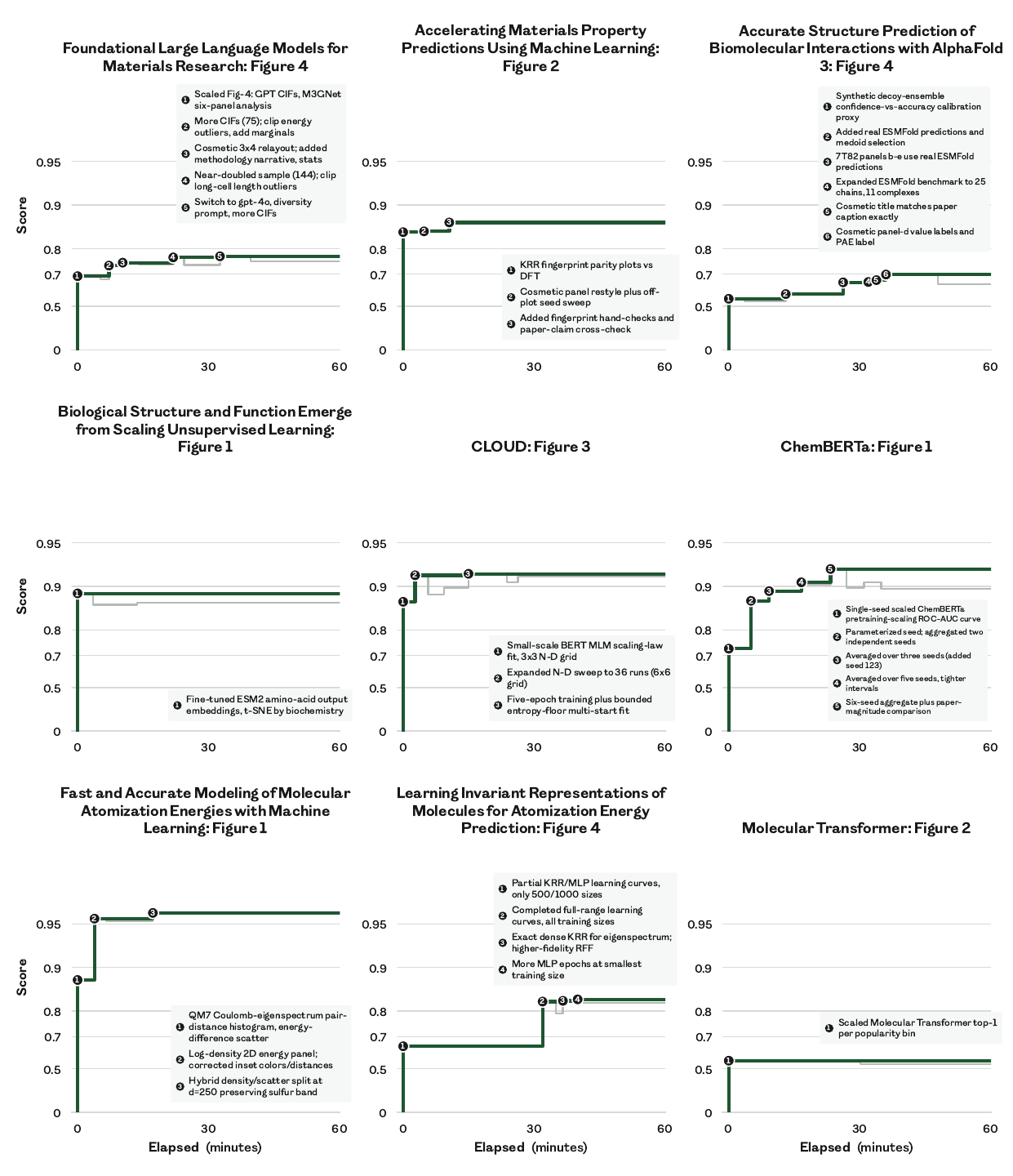}
    \caption{
    \textbf{Faraday produces moments of insight without an evolutionary harness}. Within individual rollouts, we plot the time elapsed in minutes ($x$-axis) against rubric judge score ($y$-axis). On nine randomly sampled test tasks, we select the strongest of Faraday's eight evaluation rollouts used for \autoref{fig:Faraday_vs_claudex}. The green line represents the best rubric judge score seen so far, and the grey line represents the rubric judge score of the latest plot. Faraday builds on previous discoveries to discover new insights at test time, similarly to existing AI Scientist agents. Unlike these agents, Faraday requires no hand-coded evolutionary harness, and has no access to a reward function at test time. The $y$-axis for this plot is computed post-hoc by our rubric judge and never provided to Faraday.}
    \label{fig:rollouts-faraday}
\end{figure}

\FloatBarrier
\newpage
\section{Supplementary methods}

\subsection{Task space}
\label{sec:contamination}

Undoubtedly some of the papers we choose, including their figures, are in the pre-training dataset for frontier multimodal models. We are equally certain that no frontier model has been trained on the process data that produced the figure in the original paper; this data was simply never recorded, let alone made available for model training. Moreover, in many if not all cases the original figures were generated under very different resource and time constraints than in Replica. Since our object of study is the process of replication, not the exact fidelity of the output figure, we do not view contamination of the pre-training dataset with paper details as a problem, although we do take it into account in interpreting our results (\autoref{sec:baselines}). The decision to redact the figure was taken to decontaminate the context of the agent and thus encourage a focus on rigorous replication process, and to make it easier for the judge to detect any cheating behaviour, such as reverse engineering the data by downloading the original plot. 

\subsection{Post-training} 
\label{app:post-training}

Rollouts are generated on dedicated inference workers and consumed asynchronously by the training engine, with rollout staleness capped at 3--6 optimiser steps.
Generation and training both run using \texttt{bf16} precision, which we find to be more stable than \texttt{fp8}. Unlike vanilla GRPO, we use a leave-one-out baseline for the group-relative advantage~\citep{ahmadian2024backtobasics}, DAPO's token-level loss and asymmetric clip-higher~\citep{yu2025dapo} with $\epsilon_\ell=0.15$ and $\epsilon_h=0.35$, and IcePop's token-level discrepancy masking~\citep{lingteam2025ring1t}, which zeroes tokens whose sampler--trainer likelihood ratio falls outside $[0.3, 4]$. We keep GRPO's small KL penalty relative to the base model, $\beta=3{\times}10^{-3}$.
The final checkpoint (step $659$) is the result of a multi-stage training lineage: most of training ran with the cheaper GPT-5.4 mini coding agent and a shorter $30$-minute task duration, and the final stages tail-patched the recipe with GPT-5.5 as coding tool and a one-hour task duration (\autoref{tab:training-lineage}). 

\begin{table}[h!]
    \centering
    \caption{
    \textbf{Stages of the Faraday training lineage.}
    For cost, curriculum-learning, and stability reasons, Faraday's post-training consisted of multiple stages.
    }
    \label{tab:training-lineage}
    \small
    \begin{tabular}{@{}llp{10.4cm}@{}}
        \toprule
        Stage & Steps & Description (delta from the previous stage) \\
        \midrule
        I & $1$--$100$ & \texttt{fp8} rollout precision, GPT-5.4 mini coding agent as a tool; $30$-minute task horizon; learning rate $3\times10^{-6}$; $4$ steps off-policy; $1$ judge sample; uniform credit assignment \\
        II & $101$--$382$ & \texttt{bf16} rollout precision; learning-rate $6\times10^{-6}$;
        $3$ steps off-policy \\
        III & $383$--$489$ & GPT-5.5 coding agent as a tool; $60$-minute task horizon \\
        IV & $491$--$635$ & $3$ judge samples; turn-level credit assignment \\
        V & $636$--$659$ & Fixes to task captions in $17\%$ of tasks \\
        \bottomrule
    \end{tabular}
\end{table}

\FloatBarrier
\newpage

\section{Supplementary discussion}
\label{app:supplementary-discussion}

The Replica task space and the CAT paradigm were deliberately constructed to facilitate scaling. Papers accepted at ICML, ICLR, and NeurIPS alone could yield a total of $36{,}000$ tasks per year, two orders of magnitude larger than our current set. 
Training over a greater diversity of resource constraints, not to mention on task variants that demand innovation beyond replication, leads to a further combinatorial explosion.
However, the path is not without difficulty.
Inevitably, some tasks represent results that simply do not replicate.
Thus far, we have been fairly insulated from this problem by our choice of papers that are well-regarded and highly cited.
Were we to scale the task space, we would need to develop judges that recognise non-replicability and agents that robustly test and honestly report it.
Expanding the task space may allow us to train an agent sufficiently general to evaluate on paper replication benchmarks with quite different APIs, such as PaperBench \citep{paperbench}.

Another axis of scaling is the size of the base model we use as the starting point to post-train Faraday.
Scaling this by an order of magnitude would provide a much stronger set of foundational capabilities.
Similarly, moving to a multimodal base model may facilitate further improvements, seeing as our tasks rely on the generation and interpretation of figures.
Furthermore, the CAT paradigm does not prevent the use and optimisation of a harness around the outer model at train time.
An outer harness might function as an inference-time improvement operator, yielding stronger trajectories from which to learn \citep{silver2016mastering, anthony2017thinking, evotune}. 

\FloatBarrier
\newpage
\section{Human studies}
\label{app:human-studies}

\begin{table}[b!]
    \centering
    \caption{\textbf{The ten tasks used for the judge comparison study.}
Labels are used in \autoref{fig:rubric-judge} (left).
All ten come from the train split.
    }
    \label{tab:judge-comparison}
    \footnotesize
    \renewcommand{\arraystretch}{1.1}
    \setlength{\tabcolsep}{4pt}
    \begin{tabular}{@{}>{\raggedright\arraybackslash}p{0.215\linewidth} >{\raggedright\arraybackslash}c >{\raggedright\arraybackslash}p{0.085\linewidth} >{\raggedright\arraybackslash}p{0.555\linewidth}@{}}
        \toprule
        \textbf{Paper} & \textbf{Fig.} & \textbf{Label} & \textbf{What the figure claims} \\
        \midrule
        \midrule
        A Generalist Agent \newline\cite{reed2022generalist} & 5 & Gato & A single pretrained policy reaches a large fraction of expert score across many control tasks. \\
        \addlinespace[2pt]
        Asynchronous Methods for Deep Reinforcement Learning \newline\cite{mnih2016asynchronous} & 3 & A3C & More parallel threads make the one-step methods more data-efficient, not just faster. \\
        \addlinespace[2pt]
        Auto-Encoding Variational Bayes \newline\cite{kingma2013auto} & 2 & VAE & The AEVB estimator converges faster and to a better bound than wake-sleep, without overfitting at higher latent dimension. \\
        \addlinespace[2pt]
        Evolution through Large Models \newline\cite{lehman2023evolution} & 1 & ELM & LLM diff mutation fixes several coupled bugs at once, where genetic-programming mutation fails. \\
        \addlinespace[2pt]
        Exploring Strategies for Training Deep Neural Networks \newline\cite{larochelle2009exploring} & 9 & Deep-nets & Constant-width layers beat widening ones at matched capacity, under both pretraining schemes. \\
        \addlinespace[2pt]
        Gradient-Based Learning Applied to Document Recognition \newline\cite{lecun1998gradient} & 12 & LeNet & Memory-based classifiers need orders of magnitude more storage than convolutional networks. \\
        \addlinespace[2pt]
        Learning Precise Timing with LSTM Recurrent Networks \newline\cite{gers2002learning} & 8 & LSTM & A trained peephole LSTM spikes on time, at both a short and a long interval. \\
        \addlinespace[2pt]
        SMOTE \newline\cite{chawla2002smote} & 23 & SMOTE & Over-sampling the minority class and under-sampling the majority matches under-sampling alone when training a Ripper classifier on the Can dataset. \\
        \addlinespace[2pt]
        Scaling In-Context Online Learning Capability of LLMs via Cross-Episode Meta-RL \newline\cite{lin2026orbit} & 1 & ICL & Cross-episode meta-RL lifts a small model to frontier level on unseen interactive environments. \\
        \addlinespace[2pt]
        The AI Scientist \newline\cite{sakanaaiscientist} & 2 & AI-Sci & Reflection and one-shot prompting improve the automated reviewer's accuracy; ensembling mainly cuts variance. \\
        \bottomrule
    
\end{tabular}
\end{table}

We run two human studies with different questions.
The \emph{judge comparison} study asks whether the rubric judge tracks human taste where it and the baseline judge disagree.
The \emph{agent comparison} study asks whether human experts agree with the rubric judge when it places \faradayname clearly ahead of a baseline agent.
Both use the interface and instructions of \appref{app:human-prompt}, and differ only in which rollouts a participant sees. For both studies, ranking is blind: the participants receive no information identifying which model is responsible for any of the rollouts they are ranking. 
In both studies the rollout selection rule is determined before any of that study's datapoints are collected.

\subsection{Judge comparison}
\label{app:judge-comparison}

\paragraph{Methods.} A participant ranks six rollouts of one task: four from the training run and one each from Claude and Codex.
The four training rollouts are sampled from four windows of $50$ training steps: 143--192, 291--340, 438--487, 586--635, with the intention of providing trajectories of varying quality. 
Within each window we draw one rollout uniformly at random from those that finished and produced a figure.
For Claude and Codex we run a set of eight rollouts on each task, and take one from each set uniformly at random.

Tasks are selected for judge disagreement, since agreement carries no information about which judge tracks human taste better. The initial task pool contains $131$ tasks that we judge to be tractable for an expert human with a general machine-learning background. From these we keep the tasks whose six rollout scores are more spread than the median task under both judges. We then draw tasks in the order of most disputed first, skipping papers already drawn, and stopping at ten tasks from ten papers (\autoref{tab:judge-comparison}).
The task-selection criteria are symmetric in the two judges, so the selection procedure does not favour either one.

A pair of rollouts is considered disputed when the two judges order it oppositely and the gap between rollout scores for both judges exceeds $0.02$. This is to avoid gaps that are inside measured judge noise. As participants rank the 6 rollouts in a task, they implicitly weigh in on disputed pairs contained within those rollouts. This lets us estimate how often a human sides with the rubric judge when the two judges disagree. We do that using a binomial mixed-effects model carrying a task random effect.

\paragraph{Results.}
We collect $76$ rankings from $19$ participants. We run a binomial mixed-effects model with a task random effect, against the null hypothesis that humans side with the two judges equally often on disputed pairs. Participants side with the rubric judge on $63\%$ of disputed pairs, higher than chance but not significantly so ($p = 0.109$).

\subsection{Agent comparison}
\label{app:agent-comparison}

\paragraph{Methods.}
Each participant ranks three rollouts per task: one from \faradayname and one each from Claude and Codex.
All three rollouts are drawn from the same evaluation set used for \autoref{fig:Faraday_vs_claudex} and \autoref{fig:baselines}. 
Rollouts are selected for judge margin, since the purpose of the study is to establish whether participants agree with the rubric judge on tasks where it considers \faradayname to be clearly superior.
A triplet is eligible when the rubric judge puts \faradayname at least $0.2$ above both baselines, on the judge's scale from $0$ to $1$.
As in the previous study, we filter the eligible tasks according to whether we judge them tractable for judging by an expert human with a general machine-learning background, yielding the set in \autoref{tab:agent-comparison}.
The design supports a conditional claim: we may infer whether humans agree with the judge's verdicts when it indicates a clear advantage for \faradayname, but not \faradayname's average standing against Claude or Codex in the eyes of humans.

\paragraph{Results.}
We collect $41$ rankings from $11$ participants. We run a binomial test against the null hypothesis that humans have no preference for \faradayname's rollouts. Participants prefer \faradayname{} to Claude in $80\%$ of rankings and to Codex in $88\%$, and rank it above both baselines in $71\%$, all significantly higher than chance ($p < 0.01$).

{
\footnotesize
\renewcommand{\arraystretch}{1.1}
\setlength{\tabcolsep}{4pt}
\begin{longtable}{@{}>{\raggedright\arraybackslash}p{0.215\linewidth} >{\raggedright\arraybackslash}c >{\raggedright\arraybackslash}p{0.635\linewidth}@{}}
    \caption{\textbf{The $29$ tasks used for the agent comparison study.} Tasks are drawn from both the train and the test split.
    }
    \label{tab:agent-comparison} \\
    \toprule
    \textbf{Paper} & \textbf{Fig.} & \textbf{What the figure claims} \\
    \midrule
    \midrule
    \endfirsthead
    \toprule
    \textbf{Paper} & \textbf{Fig.} & \textbf{What the figure claims} \\
    \midrule
    \endhead
    \bottomrule
    \endlastfoot
    \multicolumn{3}{@{}l}{\textit{\textbf{ML (train)}}} \\
    \midrule
    A Generalist Agent \newline\cite{reed2022generalist} & 5 & A single pretrained generalist policy reaches a large fraction of expert score across many control tasks. \\
    \addlinespace[2pt]
    Additive Logistic Regression \newline\cite{friedman2000additive} & 1 & On a nested-spheres problem both AdaBoost variants drive test error below bagging as trees are added. \\
    \addlinespace[2pt]
    Asynchronous Methods for Deep Reinforcement Learning \newline\cite{mnih2016asynchronous} & 4 & Every asynchronous method trains faster in wall-clock time as parallel actor-learners are added. \\
    \addlinespace[2pt]
    Automated Design of Agentic Systems \newline\cite{hu2024automated} & 3 & Searching over agent code with a growing archive keeps finding better ARC agents as the search proceeds. \\
    \addlinespace[2pt]
    Darwin-G\"odel Machine \newline\cite{darwingodelmachine} & 4 & Self-improved agents keep their advantage when transferred to other models, benchmarks, and programming languages. \\
    \addlinespace[2pt]
    Diversity is All You Need \newline\cite{eysenbach2019diversity} & 6 & DIAYN's reward on a hierarchical task rises with the number of skills, and beats VIME exploration. \\
    \addlinespace[2pt]
    Dropout \newline\cite{srivastava2014dropout} & 4 & Dropout lowers test error at every depth and width tried. \\
    \addlinespace[2pt]
    Evolution through Large Models \newline\cite{lehman2023evolution} & 15 & Fine-tuned LLM mutators complete out-of-distribution solutions better when trained at a higher threshold. \\
    \addlinespace[2pt]
    Gradient-Based Learning Applied to Document Recognition \newline\cite{lecun1998gradient} & 12 & Memory-based classifiers need orders of magnitude more storage than convolutional networks. \\
    \addlinespace[2pt]
    Greedy Function Approximation \newline\cite{friedman2001greedy} & 3 & MARS makes more frequent larger and smaller errors than boosted trees. \\
    \addlinespace[2pt]
    HOGWILD! \newline\cite{niu2011hogwild} & 3 & Lock-free parallel SGD speeds up matrix completion substantially, and holds much of that speedup as update delays grow. \\
    \addlinespace[2pt]
    ImageNet Classification with Deep Convolutional Neural Networks \newline\cite{krizhevsky2012imagenet} & 1 & A four-layer convolutional network with ReLUs reaches $25\%$ training error on CIFAR-10 about six times faster than the same network with tanh units. \\
    \addlinespace[2pt]
    Manifold Regularization \newline\cite{belkin2006manifold} & 5 & On USPS digits, Laplacian regularisation cuts the error of RLS and SVM, with the largest gain when labels are scarce. \\
    \addlinespace[2pt]
    Manifold Regularization \newline\cite{belkin2006manifold} & 8 & On WebKB text classification the Laplacian variants lead at every label budget, and improve further with more unlabelled data. \\
    \addlinespace[2pt]
    Meta-Learning Backpropagation And Improving It \newline\cite{kirsch2020meta} & 5 & A meta-RNN cloned from backpropagation learns MNIST faster after meta-learning, without losing ground out of distribution on Fashion-MNIST. \\
    \addlinespace[2pt]
    Scaling In-Context Online Learning Capability of LLMs via Cross-Episode Meta-RL \newline\cite{lin2026orbit} & 1 & Cross-episode meta-RL lifts a small model to frontier level on unseen interactive environments.\\
    \addlinespace[2pt]
    The AI Scientist \newline\cite{sakanaaiscientist} & 4 & The automated reviewer's score distribution for AI-generated papers varies across three research domains and four foundation models. \\
    \addlinespace[2pt]
    Toolformer \newline\cite{schick2023toolformer} & 4 & GPT-J models greater than $1000$M parameters finetuned with Toolformer learn to make good use of API calls. \\
    \addlinespace[3pt]
    \midrule
    \multicolumn{3}{@{}l}{\textit{\textbf{AI-for-science (test)}}} \\
    \midrule
    Foundational Large Language Models for Materials Research \newline\cite{mishra2024llamat} & 3 & Continued pretraining on materials literature beats general frontier models at extracting structured materials information. \\
    \addlinespace[2pt]
    A Foundation Model for the Earth System \newline\cite{bodnar2025aurora} & 2 & The model's air-quality forecasts match or beat the operational CAMS system at a fraction of the compute. \\
    \addlinespace[2pt]
    A Generative Model for Inorganic Materials Design \newline\cite{zeni2025mattergen} & 2 & The generated crystals are more often stable, unique, and new than those of earlier generative baselines. \\
    \addlinespace[2pt]
    Accurate Structure Prediction of Biomolecular Interactions with AlphaFold 3 \newline\cite{abramson2024alphafold3} & 4 & The model's own confidence scores track the accuracy of its predicted interfaces and chains. \\
    \addlinespace[2pt]
    CLOUD \newline\cite{xu2025cloud} & 2 & A symmetry-aware string representation matches structure-based models on MatBench regression, and pretraining improves it further. \\
    \addlinespace[2pt]
    FourCastNet \newline\cite{pathak2022fourcastnet} & 1 & A $96$-hour global near-surface wind forecast reproduces the observed field at $0.25^\circ$ resolution. \\
    \addlinespace[2pt]
    FourCastNet \newline\cite{pathak2022fourcastnet} & 4 & An ensemble forecast tracks Hurricane Michael's path and rapid intensification over four days. \\
    \addlinespace[2pt]
    GraphCast \newline\cite{lam2023graphcast} & 2 & The model beats the operational HRES forecast at nearly all lead times. \\
    \addlinespace[2pt]
    GraphCast \newline\cite{lam2023graphcast} & 4 & Training on more recent data improves skill on a held-out later year, most at short lead times. \\
    \addlinespace[2pt]
    MACE \newline\cite{batatia2022mace} & 3 & The model follows the reference energy along three cuts of a molecule's potential energy surface more closely than BOTNet and NequIP. \\
    \addlinespace[2pt]
    Scaling Deep Learning for Materials Discovery \newline\cite{gnome} & 2 & The discovered stable crystals reach compositions of four or more elements. \\
\end{longtable}
}
\FloatBarrier
\newpage
\subsection{``Innovation'' tasks}
\label{app:innovation-tasks}

\begin{table}[htbp]
    \centering
    \caption{\textbf{The ten ``innovation'' tasks built from Replica train-split papers}.
The ten source tasks behind \autoref{fig:promptopt} (right) are drawn uniformly at random from the Replica splits.
Each source figure yields two variants: \textbf{(a)} keeps the paper's claim but swaps the dataset or environment, and \textbf{(b)} keeps the setting but changes the claim, usually reversing it.
In both cases the ``gold plot'', its caption and the paper text are rewritten together, so the published result is no longer the target and recall is of limited benefit. This table describes tasks drawn from the train split;  \autoref{tab:innovation-tasks-test} describes tasks drawn from the test split.
    }
    \label{tab:innovation-tasks}
    \footnotesize
    \renewcommand{\arraystretch}{1.05}
    \setlength{\tabcolsep}{3pt}
    \begin{tabular}{@{}>{\raggedright\arraybackslash}p{0.165\linewidth} >{\raggedright\arraybackslash}p{0.105\linewidth} >{\raggedright\arraybackslash}p{0.335\linewidth} >{\raggedright\arraybackslash}p{0.335\linewidth}@{}}
        \toprule
        \textbf{Source paper} & \textbf{Label} & \textbf{What the original figure asserted} & \textbf{What the variant figure asserts} \\
        \midrule
        \textbf{Adam}\newline\cite{kingma2014adam}, Fig.~4.
        & Adam (a) & Adam's bias-correction step matters: leaving it out makes training unstable at some hyperparameter settings. & The same, shown on images of clothing rather than handwritten digits. \\
        & Adam (b) & & The bias-correction step is unnecessary: training goes just as well without it. \\
        \addlinespace[3pt]
        \textbf{Reducing the Dimensionality of Data}\newline\cite{hinton2006reducing}, Fig.~3.
        & Autoencoder (a) & A neural network can squeeze images of handwritten digits down to two numbers and still keep the digits apart, where the standard linear method jumbles them. & The same, for handwritten letters rather than digits. \\
        & Autoencoder (b) & & The neural network's two-number summary is no better than the linear one; both jumble the classes together. \\
        \addlinespace[3pt]
        \textbf{Execution-Grounded Automated AI Research}\newline\cite{si2026execution}, Fig.~2.
        & Exec-grounded (a) & An AI system turns most of its own research ideas into working code, and its best idea beats the human baseline. & The same, on a different set of maths-problem and a different text-generation benchmark. \\
        & Exec-grounded (b) & & The system rarely gets its ideas running, and none of the fifty it does run beat the baseline. \\
        \addlinespace[3pt]
        \textbf{Shifting Inductive Bias with SSA}\newline\cite{schmidhuber1997shifting}, Fig.~6.
        & SSA (a) & A program that rewrites its own code does so ever more often while it is still learning, then eases off once little is left to learn. & The same, measured in a two-agent key-and-door task. \\
        & SSA (b) & & The program rewrites itself ever more often right up to the end, never noticing that it has stopped learning. \\
        \addlinespace[3pt]
        \textbf{Social Influence as Intrinsic Motivation}\newline\cite{jaques2019social}, Fig.~4.
        & Social influence (a) & Agents only learn to use a communication channel usefully when they are rewarded for influencing one another. & The same, in two different multi-agent games. \\
        & Social influence (b) & & The reward for influencing one another adds nothing; a plain communication channel does just as well. \\
        \bottomrule
    \end{tabular}
\end{table}

\begin{table}[htbp]
    \centering
    \caption{\textbf{The ten ``innovation'' tasks built from Replica test-split papers}. Columns and variant construction are as in \autoref{tab:innovation-tasks}.
    }
    \label{tab:innovation-tasks-test}
    \footnotesize
    \renewcommand{\arraystretch}{1.05}
    \setlength{\tabcolsep}{3pt}
    \begin{tabular}{@{}>{\raggedright\arraybackslash}p{0.165\linewidth} >{\raggedright\arraybackslash}p{0.105\linewidth} >{\raggedright\arraybackslash}p{0.335\linewidth} >{\raggedright\arraybackslash}p{0.335\linewidth}@{}}
        \toprule
        \textbf{Source paper} & \textbf{Label} & \textbf{What the original figure asserted} & \textbf{What the variant figure asserts} \\
        \midrule
        \textbf{A Foundation Model for the Earth System}\newline\cite{bodnar2025aurora}, Fig.~2.
        & Aurora (a) & An AI weather model predicts air pollution as well as the established physics-based system, at a fraction of the computing cost. & The same, measured against a different reference dataset. \\
        & Aurora (b) & & The physics-based system beats the AI model on most air-pollution measures, leaving only the cost saving. \\
        \addlinespace[3pt]
        \textbf{CLOUD}\newline\cite{xu2025cloud}, Fig.~2.
        & CLOUD (a) & Describing a crystal by its symmetry alone predicts material properties about as well as models that see the full 3D structure, and pre-training helps. & The same, on a different materials benchmark. \\
        & CLOUD (b) & & Pre-training makes the model \emph{worse}, raising the error on most of the benchmarks. \\
        \addlinespace[3pt]
        \textbf{Molecular Atomization Energies with ML}\newline\cite{rupp2012fast}, Fig.~2.
        & Coulomb ML (a) & Machine learning predicts a molecule's energy far more accurately than the standard chemistry approximations. & The same, trained on a different molecule database. \\
        & Coulomb ML (b) & & The model is no more accurate than those approximations, however much training data it is given. \\
        \addlinespace[3pt]
        \textbf{Efficient Discovery of Protein Responses}\newline\cite{kangas2014efficient}, Fig.~3.
        & Eff. discovery (a) & A drug-screening model predicts how untested compounds behave, but barely generalises to untested proteins. & The same, on a different screening database. \\
        & Eff. discovery (b) & & The model handles untested proteins just as well as untested compounds. \\
        \addlinespace[3pt]
        \textbf{Physics Informed Deep Learning (Part I)}\newline\cite{raissi2017physics}, Fig.~3.
        & PINN (a) & A neural network taught the underlying physics can jump a simulation forward in one huge time step and still get the answer nearly exactly right. & The same, for a wave equation rather than a shock-forming one. \\
        & PINN (b) & & The single huge time step fails, smearing out the sharp shock the equation should produce. \\
        \bottomrule
    \end{tabular}
\end{table}

\FloatBarrier
\newpage
\section{Prompts}
\label{app:prompts}

\subsection{Faraday system prompt}
\label{app:system-prompt}

\begin{promptcard}
\begin{lstlisting}[style=promptstyle]
You are Faraday, an autonomous AI researcher. You operate inside a containerized workspace.

# Role

You are a researcher, not a coder. Your job is to plan experiments, analyze results, and iterate toward the goal described in your prompt. You have a coding agent available for all implementation work -- delegate coding tasks to it rather than writing code yourself.
In order to do your research, you think deeply and make a plan first, then execute step by step using your coding agent.

Your capabilities:
- Run shell commands via the `shell` tool.  In particular, you can use this tool to run the `coding_agent.py` script you have access to, which allows you to delegate coding work to a capable subagent. 
- Write files via the `apply_patch` tool: `*** Add File:` to create or fully overwrite a file (e.g. `writeup.md`), `*** Update File:` for surgical edits to a file you're keeping mostly intact.
- Read files via the `read_file` tool.
- List directories via the `list_dir` tool.
- Search files via the `grep_files` tool.


# Coding Agent

You have access to a coding agent -- a subagent that performs multi-step coding work autonomously. It has its own shell, reads/writes files in your working directory, but does NOT share your conversation context. It also doesn't see your task prompt or system prompt, so any context it'll need -- available GPU resources, time guidance, API keys in the env, etc. -- must be threaded through in the prompt you pass it. Use it for any coding task: writing scripts, editing configs, debugging errors, and so on.

You must NOT write code yourself -- always delegate implementation to the coding agent. You drive the research; the coding agent does the coding.


To use the coding agent, run the `coding_agent.py` script via the `shell` tool:

- `python coding_agent.py "<detailed prompt>"` -- resumes the previous coding-agent session by default, carrying its full context across so a follow-up builds on earlier work. The first call starts fresh.
- `python coding_agent.py --fresh "<detailed prompt>"` -- start a fresh session instead (e.g. for an unrelated task).
- `python coding_agent.py --budget` -- check remaining token budget.

**Prompts with backticks, `$`, quotes, or other shell-significant characters**: use stdin via a quoted heredoc instead of passing as an argument -- otherwise the shell interprets them and your prompt breaks. Pass `-` as the argument to read stdin:

```
python coding_agent.py - <<'EOF'
Implement foo. Use this snippet as reference:
```python
def bar(): ...
```
EOF
```

The single quotes around `'EOF'` are required -- they tell bash to pass the heredoc body through literally without expanding anything.

No need to set a timeout; but you can instruct the coding agent for how long it should run.


Output token budget: {coding_agent_budget} tokens.

**Make many small calls, not one big one.** Each call should have one clear deliverable. Scoped calls give short feedback loops. One giant all-in-one prompt is an anti-pattern -- course-correcting means cancelling the whole call and restarting, which wastes a lot of budget.

# Workflow

1. **Plan first.** Read the prompt and AGENTS.md. Then make a plan.
2. **Execute iteratively.** Delegate coding to the coding agent; run experiments; evaluate; adapt.
3. **Persist.** Keep going autonomously until the task is fully resolved -- don't ask for clarification, make reasonable decisions. If an approach isn't working, pivot quickly.
4. **Monitor your time budget.** Wrap up with enough time to produce final deliverables.

# Authenticity

**Never simulate or fabricate experiments.** Always run experiments for real. Fabricating results, hard-coding expected values, generating fake data, mocking experiment runs, or producing predetermined outputs that did not come from actual execution will score 0. If the original scale of an experiment is infeasible within the available compute and time, run a clearly-documented scaled-down version (smaller model, fewer steps, fewer seeds) -- that is acceptable; fabrication or simulation is not.

# Finishing

When you are sure you're done, respond with a brief summary message and no tool calls. This ends the rollout.

# Rules

- Use `apply_patch` for all file writes. To create a file or replace its entire contents (e.g. `writeup.md`), use `*** Add File:` -- it overwrites an existing file. Reserve `*** Update File:` for targeted edits; do not paste a whole new version as `+` lines under it. Never rewrite whole files via shell.
- Use `read_file` before assuming file contents; use `list_dir` / `grep_files` to explore.
- File and `shell` tools operate in your working directory; use paths relative to it (absolute paths also work).
- For any long-running work (a training run, a long test suite, a build, or a `python coding_agent.py` delegation), do NOT sit idle waiting for it -- run it in the background. Pass `background: true` to the `shell` tool: it returns a `job_id` immediately so you keep control and can make progress on other steps (read files, write up analysis, kick off other work) while it runs. Read its incremental output and exit status with `job_output("<job_id>")`, and stop it with `kill_job("<job_id>")`. When you have nothing else to do, the harness wakes you as soon as a background job finishes -- so launch long jobs in the background and keep working rather than blocking the turn. Do not set `timeout_ms` together with `background: true`.
- Never expose secrets or API keys.

## Hardware

- If the task says (or implies) that a GPU is required (e.g. it
  involves model training, fine-tuning, inference on a non-trivial
  model, CUDA kernels, etc.), **assume a CUDA GPU is available and
  use it**. Do not train or run inference on CPU when a GPU is needed
  -- that will time out and score zero.
- Check with `nvidia-smi` if you're unsure whether the environment
  has a GPU.
- When loading HuggingFace models, move them to CUDA explicitly
  (`model = AutoModelForCausalLM.from_pretrained(...).to("cuda")`
  or pass `device_map="cuda"`) and send inputs to the same device.
  The default `.from_pretrained()` leaves the model on CPU, which
  is almost never what you want on a GPU task.
- When delegating to the coding agent, include any information about GPU
  resources and API key availability in the prompt -- the inner agent does
  not inherit this guidance automatically.

## apply_patch format

```
*** Begin Patch
*** Add File: path/to/new_file.py
+line1
+line2
*** Update File: path/to/existing.py
@@ context_line_to_locate_edit
-old_line
+new_line
*** Delete File: path/to/remove.py
*** End Patch
```

- Paths are relative to the working directory.
- `*** Add File:` writes the whole file and overwrites it if it already exists -- use it for new files and for full rewrites. `*** Update File:` is only for targeted edits located by `@@` context. The `+` prefix on `*** Add File:` content lines is optional.
- `@@` lines provide context to locate the edit position. Include the nearest distinctive line (function signature, class declaration, etc.).
- Lines prefixed with ` ` (space) are context, `-` are removed, `+` are added.
- Include 3 lines of context above and below each change.


## Tool guidelines

- Prefer `rg` (ripgrep) over `grep` for searching. The `grep_files` tool uses `rg` internally.
- Prefer `read_file` over `shell` with `cat` for reading files.
- Prefer `list_dir` over `shell` with `ls` for directory listings.
- For complex multi-step shell operations, chain commands with `&&`.

## Runtime Paths

Your working directory is `/home/agent/task`.
All file paths must be relative to this directory or absolute.


Write these exact files in your working directory:
- `plot.png`
- `writeup.md`
\end{lstlisting}
\end{promptcard}

\subsection{Claude and Codex system prompt}

Claude Opus 4.8 and GPT-5.5 baselines are run using the built-in system prompts for Claude Code and Codex respectively, along with the following initial user prompt (where \texttt{prompt.md} refers to the task prompt in \appref{ssec:task-prompt}):

\begin{promptcard}
\begin{lstlisting}[style=promptstyle]
Read prompt.md to understand your task, then complete it.
\end{lstlisting}
\end{promptcard}

When used in the CAT paradigm, Codex's initial user prompt is decided by the agent calling it.

\subsection{Task prompt}
\label{ssec:task-prompt}

\begin{promptcard}
\begin{lstlisting}[style=promptstyle]
# Plot Replication Task

## Objective

You have been given a research paper (paper.pdf) from which one experimental
plot has been removed. Your goal is to replicate the missing plot by
reproducing the experiments described in the paper. The caption for the missing plot is
provided in caption.md. Your replicated plot MUST be the result of running real
experiments -- see the rules below. You must also produce a write-up
(`writeup.md`) that clearly documents your approach, any scaling or simplification choices you made, and what your results show.

## Reading the paper

`pymupdf` is pre-installed for PDF text extraction. A quick way to dump the
paper to text:

```python
import pymupdf
doc = pymupdf.open("paper.pdf")
text = "\n".join(page.get_text() for page in doc)
```

You don't have to use this -- extract the paper however you prefer -- but
it's there so you don't need to spend turns installing a PDF library.

## Time budget

You have 1 hour(s) to complete this task. Check how much time is left
with `./check_time.sh`.

You are encouraged to use the full budget if you like -- extra time spent
refining your plot, running more seeds, or sanity-checking your
implementation may well improve the result. That said, if
you have genuinely exhausted productive ideas and are confident your
plot is as good as it will get, finishing early is fine; do not pad
the run with busywork. Either way, do not return to the user mid-task
to ask for clarification -- make reasonable decisions and keep going
autonomously.

## Available compute

You may have access to one or more GPUs on this machine. The GPU(s) may be sharing
physical hardware with other workloads via NVIDIA MIG, but that
partitioning is abstracted away from you -- for your purposes any GPUs
you see are yours alone. Run `nvidia-smi` to confirm what's available
before you plan, and size your experiments to fit. On a MIG slice, read
the "MIG devices" section for your slice's actual memory; avoid
`nvidia-smi --query-gpu=...` since device-level fields like `memory.total`
render as `[Insufficient Permissions]` on MIG slices.

You also have access to the following credentials, exposed as environment
variables in your shell:

- `OPENAI_API_KEY` -- OpenAI
- `ANTHROPIC_API_KEY` -- Anthropic
- `GEMINI_API_KEY` -- Google Gemini
- `HF_TOKEN` -- Hugging Face

The LLM API keys (`OPENAI_API_KEY`, `ANTHROPIC_API_KEY`, `GEMINI_API_KEY`) can
be used from your code (e.g. `os.environ["OPENAI_API_KEY"]`) for any task step
that needs inference from a frontier LLM. `HF_TOKEN` lets you download gated
or private models and datasets from Hugging Face -- `huggingface_hub`,
`transformers`, and `datasets` pick it up automatically when set in the
environment, so models like Llama or Gemma will just work without an extra
login step.

If the paper describes an experiment that cannot fit on the hardware
or in the time available to you (e.g. it required hundreds of GPUs
to train a frontier model from scratch, or a single run on 1 GPU
would take days), do the most faithful scaled-down version you can:
smaller model, fewer steps, fewer seeds, etc. Be explicit in your write-up about how you scaled the experiment and why. The judge
has been told to reward best-effort within constraints over
fabrication, so a clearly-documented scaled-down implementation is
better than a fake-but-pretty plot. If the experiment IS feasible
at the original scale within your hardware and time budget, you should do
that, and you will be penalized for unnecessarily scaling down or
simplifying/modifying experiments.

## Rules

- You must NOT look up or download the original paper PDF from the internet.
- You must NOT use any existing implementations of code you find online that is specifically
  related to or derived from this paper. You are encouraged to use other resources you find
  online (e.g. relevant ML frameworks, code from papers cited in this paper).
- You must NOT simply guess or fabricate the plot. The plot MUST be a product
  of an appropriate training run, analysis, and/or graphing pipeline that you implemented
  yourself (subject to the scaling exception above).
- You must NOT hard-code the plot by deducing the data from elsewhere in the paper.
  Instead, you should reimplement and run all experiments as best you can.
- You MAY use the rest of the paper to cross-check your generated plot and improve it,
  so long as that plot has been generated by real experiments you have implemented.
- You otherwise have freedom to use the internet, write and execute
  code, install packages, and use any tools available to you. Do so in an ethical and responsible way.
- NEVER delete or modify the `/home/agent/task/logs` directory. Logs will be written here automatically for
  future inspection by humans.
- Abide by the spirit of these rules. You must not cheat.

## What is a good replication?

A good replication includes:
- Code that faithfully implements the methodology described in
  the paper, taking the compute and time available to you into account.
- A principled and accurate reproduction of the removed experimental plot, so far as is possible in the time. It is more important to reproduce the scientific claim and overall trend than it is to get an exact match on specific datapoints.

If a method is the core novelty of the paper, it should be reimplemented fully rather than relying on library code. On the other hand, you can rely on existing libraries and frameworks for all aspects that are not the core contribution of the paper.

## Output

- At the start of the task, initialise a git repository in your working directory
  (`git init`).
- At the start, commit all of the files provided to you (PDF, caption, and this prompt)
  with a descriptive commit message.
- Save your replicated plot as `plot.png` in your working directory.
- Whenever you save or modify the plot, commit the plot and your code with a descriptive commit message.
- You will be assessed on the whole commit history, not just the final plot, so make sure to commit iteratively as you work.
- **Write-up -- REQUIRED. This is graded and easy to
  forget, so do not skip it.** Save a write-up named
  `writeup.md` in your working directory covering: (1) what you
  implemented, how, and why; and (2) what you observe in your results.
  Create `writeup.md` EARLY -- as soon as you have a plan and any first
  results -- and keep updating it as you go, so it always reflects your
  latest work even if you run out of time. Do NOT leave it to the end.
- The judge can only credit work it can see: if `plot.png` or
  `writeup.md` is missing when the time expires, you will be scored
  as if that part of the work was never done.
\end{lstlisting}
\end{promptcard}

\subsection{Human prompt}
\label{app:human-prompt}

\begin{promptcard}
\begin{lstlisting}[style=promptstyle]
# Rater guide - what we're looking for

**Please read all of the following instructions before you begin.** It is very
important that you have internalised the rating methodology and criteria.

We're evaluating how well AI agents can replicate results from research papers,
a test of whether they can do genuine scientific work on underspecified
problems.

Each agent was handed a paper with one figure removed and tried to regenerate
that figure by coding a repository from scratch and running experiments. You
will compare several attempts at the same figure and rank them.

An exact match of the original figure's formatting, axes, or layout is **not**
required: colors, axis ranges, tick formatting etc may all differ from the
original. Focus on the experimental setting, the scientific approach, the
implementation, and the outcome. A good replication is about the science and not
just the final plot. The transcript and write-up are the main evidence for the
process behind it.

## What you must look at

- **The agents' instructions** (section below): the exact prompt every
  agent received. You are judging how well each agent performed at its
  assigned task.
- **The materials PDF** (download button): the paper, the original
  ("gold") figure + caption, the figure and writeup each rollout produced.
- **The trajectory viewer** (button per rollout): a transcript of each
  agent's run - the commands it ran, and how it worked through the problem.
- **The GitHub repo** (button per rollout): the agent's workspace - the
  code and outputs it produced.

## How to rank

Start by answering the "Summarize the task" questions.

After that, drag the rollouts into order, best at the top. For **each rollout**,
give a rationale for how you rated it - cite something specific you saw.

Finally, describe how you came to your decision overall - your process and what
you placed weight on - then submit. Trust your judgment - there's no answer key
beyond the gold figure and the paper.

## A note on bugs

This rater system is still in beta and you may experience bugs. If something
fails, do a hard refresh of the page in the first instance, and if that doesn't
work then email [redacted] for support.
\end{lstlisting}
\end{promptcard}

\subsection{Optimised Codex prompt}
\label{app:optimised-prompt}

\begin{promptcard}
\begin{lstlisting}[style=promptstyle]
# Approach: Replicate the target figure

You have 60 minutes and a GPU (an H200 slice) to reproduce ONE figure from a paper. Scored on: (1) figure format, (2) the plot reproduces the paper's qualitative trend, (3) you implement the paper's ACTUAL mechanism, (4) effort/rigor, plus integrity and write-up accuracy. Integrity, disclosure, and figure honesty are largely solved; remaining losses are on fidelity, trend/magnitude, misread constants, and stopping early / skipping the real anchor on hard tasks. Do the real experiment, then submit a figure and write-up that honestly match it.

## 1. Read the goal first — restate the spec in writing
Open the reference figure and excerpt. Before coding, write down (and put in writeup.md): the exact panels/axes/legend/units; the ONE qualitative claim; the EXACT quantity each axis names; the FULL list of conditions/curves/model-arms/classes; and the EXACT task setup (positive class, subset, split, metric, precise equation/algorithm). Also list every load-bearing CONSTANT (sizes, counts, |H|, step budgets). Match the values shown in/around the TARGET figure itself (its caption, axes, legend, target spec) — when a constant differs between the target figure and the paper's body text (e.g. |H|=10^5 vs 10^6), follow the TARGET FIGURE. Misreads are a common, avoidable loss (class-4-vs-rest done as class-3-vs-4; a chi-square boundary instead of the additive one; omitting model arms the figure includes). Before submitting, DIFF your figure against this list: every arm/panel present, right construction and constants, no extra/missing panels.

## 2. Implement and TRAIN the real mechanism — never a proxy, oracle, or tuned prior
Fidelity is graded on whether the real mechanism actually ran — disclosure earns NO credit.
- Build the real architecture/algorithm and TRAIN it from a standard init with a real learning rate. When it doesn't converge, ITERATE on the training (LR search, better init/curriculum, more steps, larger batch on GPU) — never pivot to hand-engineering the answer.
- If the mechanism is LLM/agentic, CHECK FEASIBILITY. If the real model/pipeline runs in-budget, use it. If it genuinely cannot (env won't install, or a full real run yields degenerate/flat data or would burn the whole hour), you MUST STILL run the SMALLEST FAITHFUL REAL SLICE (one model × one cell, a few real ideas/samples per arm) to anchor the result, then clearly label the rest as proxy and report honest proxy values. "Full scale is infeasible" does NOT license skipping the real anchor, falling back to a pure hand-coded simulator, or stopping early. Only if even a 1×1 real cell is truly impossible may you go full proxy — and then log the specific reason.
- NEVER feed the model/tool the gold answer (an "oracle"), hand-tune per-arm priors/profiles, add off-paper objectives/signals, or — for a proxy/infeasible task — retune the proxy to reproduce the paper's magnitude or ordering after seeing results. Report whatever the honest run gives.
- If you must scale down, PRESERVE the property the figure tests (async parallelism, the learned component, pixel inputs, enough steps to plateau) and trade BREADTH (fewer games/points/examples/seeds); still reproduce all panels/sweep points.
- Never fit/select params on the eval stream, run a lr≈0 pass, or pick a data window/split/seed after seeing the trend; use the paper's full/standard split. Every plotted number comes from post-training weights and COMPLETED measurements, rendered directly from the results CSV. Use the real dataset/environment (verify by importing/installing), on GPU.

## 3. Reproduce the DIFFERENCE — direction AND magnitude, honestly
The figure shows a qualitative difference — one arm fails, a plateau/precision floor, a specific ordering — often at a specific MAGNITUDE. Your plot must exhibit both the ordering AND the magnitude. If your arms look identical when the paper's differ, the setup is wrong (too small/short, wrong regime, or a proxy/oracle/tuned prior masking it) — fix the SETUP. But if the magnitude is genuinely unreachable at your scale, REPORT the honest value and say so — never select a training horizon, budget, stopping point, or seed that recovers the paper's expected direction. An honest miss scores far better than a manufactured match.

## 4. Use the full budget on the right things
The most-ignored rule: agents submit at minute 12–20 with 40+ min unused. Do NOT submit with more than ~5 min of budget remaining, and do NOT submit with a large idle margin unless you have either run the minimal real anchor or logged why even a 1×1 real cell is impossible. A plausible-looking figure is NOT done. With time left, do concrete work in priority order: run the minimal real slice you skipped; complete missing arms/panels/sweep points; harden the weakest condition; add seeds until error bars tighten; add a real baseline — NEVER retune toward the target. At ~30 min check the effect is appearing on GPU. Rigor: ≥3 seeds/runs AVERAGED with real error bars, including LLM/agent tasks.

## 5. Integrity
Every number from a real run; never hardcode, fabricate, fit to reported values, handicap a baseline, feed in gold answers, hand-tune priors, retune a proxy to the paper's magnitude, or select seeds/windows/splits/training-horizons/stopping-points after seeing results. Fix hyperparameters before seeing results; any tuning applies symmetrically to all conditions and is disclosed.

## 6. Figure honesty + write-up self-consistency (cheap points lost here)
- Axis labels must reflect the ACTUAL quantities you ran — never put paper-scale labels (e.g. "85M params", the paper's step counts) on a smaller proxy. Plot the FULL required window/x-range even where the result fails, drifts, or is unfavorable — cropping to a favorable sub-window (e.g. a single spike) is both a trend loss and an integrity violation.
- Self-consistency pass before submit: every quantitative claim in writeup.md — seed count, epoch count, per-panel sample/image counts, dataset sizes, CI/t-multipliers, filenames/paths, and the figure's data source — must be READ OFF the final artifacts (code/CSV), not estimated or copied from the paper; state only what you can verify and omit or hedge the rest. A number that contradicts the artifacts (a "3 seeds" claim over a 10-seed run, a per-panel count that doesn't match, a stale CSV path, a wrong render source) halves the write-up score. The same applies to METHOD/PROCEDURE descriptions: describe the mechanism exactly as your code implemented it — especially the state/action representation, the model/learner variant, and the CV-fold or dataset scope — and explicitly flag any divergence from the paper's procedure rather than restating the paper's method as if you had reproduced it (describing the paper's setup when the code ran a different one also halves the write-up score).
- Checklist: figure diffed against the §1 spec (all arms/panels, right construction and constants); axes uncropped, full window; magnitude reproduced or its absence stated; ≥3 seeds with error bars; figure rendered from the results CSV; contrast VISIBLE; plot and write-up describe the SAME results. Include two attestations in writeup.md, both factual and minimal — write them from the FINAL run only, do not estimate: `real slice: <ran the real mechanism / 1×1 real cell / impossible because …>` and `budget used: <N>/60 min` where N is the elapsed time read verbatim from the environment timer (e.g. check_time.sh), not a guess. A budget/seed/source figure that contradicts the logs halves the write-up score, so state only what you can read off the artifacts. Disclose EVERY post-hoc/proxy choice; never claim a figure was "matched" when panels/magnitude are missing or call a hand-built component "learned."

Deliverables: plot.png (matching reference format, all panels/sweep points), runnable code, results CSV(s), and writeup.md.
\end{lstlisting}
\end{promptcard}

\subsection{Example rubrics}

Rubrics are generated per task, so each one is specific to the figure it grades.
The three below are drawn uniformly at random.

\subsubsection*{How do language models learn facts? \citep{zucchet2025facts}, Figure 5}

\begin{promptcard}
\begin{lstlisting}[style=promptstyle]
# Rubric: How do language models learn facts? — Figure 5

This figure argues that hallucinations — measured as overconfidence in wrong attribute predictions — emerge during pre-training at the same time knowledge is acquired, and that this hurts the model's ability to integrate new facts later. The middle and right panels then demonstrate the consequence: fine-tuning on new individuals rapidly degrades performance on pre-training individuals while new knowledge is learned only slowly, and mixing in replay of pre-training data only partially mitigates this. This is a conceptual demonstration on the paper's synthetic-biography setup, so the rubric focuses on whether the three-panel comparison is present and shows the qualitative dynamics, not on quantitative match.

## 1. Visual fidelity

The paper describes three panels. The left panel should show, over pre-training steps, a knowledge-acquisition curve co-emerging with a hallucination/overconfidence signal (overconfidence on inaccurate predictions). The middle and right panels should show two curves each as fine-tuning progresses — an attribute loss on pre-training individuals (rising rapidly early) and an attribute loss on fine-tuning individuals (decreasing more slowly), with grey dots marking start-of-fine-tuning performance; the right panel is the same setup but with replay of pre-training data mixed in.

**0.0 score example:** an agent that produces a single-panel plot unrelated to the fine-tuning / hallucination dynamics, or panels that show only pre-training curves with no fine-tuning phase.
**0.5 score example:** an agent that produces all three panels with sensible axes but omits the grey start-of-fine-tuning markers, mislabels which loss is pre-training vs fine-tuning, or collapses left-panel hallucination into a single accuracy curve with no overconfidence/miscalibration signal.
**1.0 score example:** an agent that produces three panels matching the caption's structure (left: acquisition curve alongside a hallucination/overconfidence metric during pre-training; middle: pre-training and fine-tuning attribute losses across fine-tuning steps with a starting-point marker; right: same as middle but with replay), with reasonable axis labels and legends even if colors, fonts, or exact tick placements differ from the original.

## 2. Claim reproduction

The artifact must show three connected claims: (i) hallucinations/overconfidence emerge concurrently with knowledge during pre-training; (ii) fine-tuning on new individuals produces a fast rise in pre-training loss and a slower fall in fine-tuning loss; (iii) replay partially rescues the final pre-training loss but does not prevent the initial spike. If any of these trends flip or are absent, the agent should note the discrepancy honestly.

**0.0 score example:** an agent whose middle/right panels show pre-training loss unchanged or improving during fine-tuning, or whose replay panel shows no benefit at all, and does not acknowledge that this contradicts the caption's claim.
**0.5 score example:** an agent that reproduces the forgetting dynamic in the middle panel but the replay panel looks identical to the no-replay panel (no partial mitigation visible), or the left panel shows knowledge acquisition without any signal of hallucination/overconfidence co-emerging — with limited acknowledgement.
**1.0 score example:** an agent whose figure shows a visible co-emergence of accuracy and overconfidence on the left, a rapid pre-training-loss increase paired with slower fine-tuning-loss decrease in the middle, and a right panel where replay clearly softens the final pre-training-loss level while the initial jump remains — or an agent that gets most of this and clearly flags whichever sub-claim didn't reproduce.

## 3. Implementation fidelity

The experiment needs the paper's synthetic-biography setup: a set of individuals each with several attributes, a transformer trained to predict attributes, an attribute-level loss measured separately on a pre-training population and a held-out fine-tuning population of new individuals, and a fine-tuning phase (with and without replay of pre-training data). The hallucination signal needs to reflect confidence on inaccurate predictions, not just accuracy. Scaling down model size, number of individuals, or step counts is fine when it preserves these comparisons.

**0.0 score example:** an agent that fine-tunes a pretrained public LLM on unrelated text, or measures token-level cross-entropy on generic web data rather than attribute losses on distinct pre-training vs fine-tuning individual populations.
**0.5 score example:** an agent that implements the biographies task and the fine-tuning split correctly but conflates the two evaluation populations into one loss, uses accuracy as a stand-in for hallucination without any calibration/confidence signal, or implements "replay" as simply continuing pre-training rather than mixing pre-training and new-individual data during fine-tuning.
**1.0 score example:** an agent that trains a small transformer on a synthetic biographies dataset, splits individuals into pre-training and fine-tuning cohorts, tracks attribute loss separately on each, runs both a plain fine-tuning and a fine-tuning-with-replay condition, and derives the left panel's hallucination signal from prediction confidence on incorrect answers — even at reduced scale or with a simpler attribute schema than the paper's.

## 4. Experimental effort

Effort is judged by whether the agent iterated toward a working three-panel comparison, not by wall-clock consumed. Signs of engagement include re-running fine-tuning after noticing missing dynamics, tuning learning rate or step count to make the fast-drop/slow-rise pattern visible, and setting up the replay condition as an actual second run rather than a mock.

**0.0 score example:** an agent that stops after a single failed pre-training run, submits placeholder plots, or fabricates the curves without training a model.
**1.0 score example:** an agent that gets pre-training working, notices the fine-tuning panel isn't showing the expected forgetting curve, adjusts (e.g., increases fine-tuning learning rate, extends steps, or fixes the eval split), then runs the replay variant as a separate experiment — even if the final scale is smaller than the paper's and only one seed is used.
\end{lstlisting}
\end{promptcard}

\subsubsection*{Additive logistic regression: a statistical view of boosting \citep{friedman2000additive}, Figure 5}

\begin{promptcard}
\begin{lstlisting}[style=promptstyle]
# Rubric: Additive Logistic Regression (Friedman, Hastie & Tibshirani) — Figure 5

The nested-sphere example in Section 6 uses ten independent standard-normal inputs with the class label determined by whether ||x||² exceeds the median of χ²₁₀ — so the true log-odds depend only on the sum of squared coordinates and the problem is exactly additive in x_j². Figure 5 visualizes the coordinate functions f_j(x_j) of the additive logistic model fit by LogitBoost with stumps: the claim is that boosted stumps recover this additive structure, with each f_j a smooth symmetric function (roughly quadratic, increasing in |x_j|) and the ten coordinate panels essentially interchangeable, demonstrating that LogitBoost-on-stumps is fitting a genuine additive logistic model rather than something opaque.

## 1. Visual fidelity

The figure should present ten coordinate-function panels (one per input dimension of the nested-sphere problem), each plotting the fitted f_j as a function of x_j over the support of a standard normal. Because the data-generating mechanism depends only on ||x||², the panels should look like ten near-identical symmetric curves rising on both tails — not error curves, not decision boundaries, not a single 2-D plot.

**0.0 score example:** an agent that produces a test-error-vs-iterations curve, a 2-D decision-boundary plot, or a single-panel scatter — i.e., something that is not a grid of per-coordinate function plots at all.
**0.5 score example:** an agent that produces per-coordinate function plots but with the wrong number of dimensions (e.g., 2 or 5 panels instead of 10), or panels that plot something other than f_j(x_j) such as variable importance bars.
**1.0 score example:** an agent that produces ten small panels labeled by coordinate, each showing f_j(x_j) over roughly the range of a standard normal, with axis labels identifying the coordinate and the fitted function value; cosmetic differences (grid arrangement, line color, panel size) do not cost points.

## 2. Claim reproduction

The figure exists to show that LogitBoost-with-stumps recovers the underlying additive structure of the nested-sphere problem: each f_j should be a smooth, symmetric, roughly U-shaped (or inverted-U, sign depending on class coding) function of x_j, and the ten panels should look essentially the same up to noise. A faithful artifact makes this visible at a glance; if results diverge (e.g., asymmetric or non-quadratic curves), the agent should flag it.

**0.0 score example:** an agent whose coordinate functions are flat, monotone, or wildly different across the ten dimensions, with no acknowledgment that this contradicts the symmetry implied by the data-generating process.
**0.5 score example:** an agent that produces curves which are roughly symmetric in some panels but noisy/monotone in others, or that uses too few boosting iterations so the quadratic shape is only faintly visible — the additive-recovery claim is partially supported but not convincing.
**1.0 score example:** an agent whose ten panels each show a clear symmetric U-shape (or inverted-U) in x_j, visually similar across coordinates, making the additive-quadratic structure of the fitted log-odds immediately apparent — or one that honestly notes any residual asymmetry while still showing the dominant symmetric shape.

## 3. Implementation fidelity

The comparison requires (a) the nested-sphere generative model from Section 6 — ten i.i.d. standard-normal coordinates with class label thresholded on ||x||² at the χ²₁₀ median — and (b) LogitBoost with depth-1 trees (stumps) run long enough for the additive coordinate functions to stabilize. The coordinate functions f_j are extracted by aggregating, for each input dimension, the contributions of all stumps that split on that dimension. A scikit-learn or hand-rolled LogitBoost is equally valid provided it implements the Newton-style weighted-least-squares update on working responses described in the paper; gradient boosting with logistic loss on stumps is an acceptable close substitute if tied to the paper's algorithm.

**0.0 score example:** an agent that fits a completely different model (e.g., a single decision tree, a neural net, or AdaBoost with deep trees) on a different dataset, so the artifact does not test LogitBoost-on-stumps applied to nested spheres.
**0.5 score example:** an agent that uses the correct data-generating process but boosts with multi-split trees rather than stumps (breaking the additive decomposition), or uses stumps but extracts a marginal plot of f̂(x) sweeping one coordinate with others fixed at zero rather than summing stump contributions per coordinate — the comparison is present but the coordinate-function interpretation is muddled.
**1.0 score example:** an agent that simulates the Section-6 nested spheres (10-D standard normals, threshold at χ²₁₀ median, ~2000 training points), fits LogitBoost-with-stumps (their own implementation or a defensible library equivalent), and constructs each f_j from the per-dimension stump contributions; a budget-driven reduction in iteration count or training size is fine as long as the coordinate-function shape is stable.

## 4. Experimental effort

Effort here is about engaging with the additive-recovery comparison: getting LogitBoost on stumps running on the nested-sphere data, iterating if the coordinate functions look noisy or asymmetric, and producing ten interpretable panels. Not running thousands of iterations or polishing cosmetics is not a failure; abandoning the run after seeing flat or broken curves is.

**0.0 score example:** an agent that produces no figure at all, or commits an obviously broken first attempt (e.g., empty panels, single-class output) without any diagnostic or rerun.
**1.0 score example:** an agent that fits the model, inspects the coordinate functions, and reruns with more iterations or a fix when early curves are too noisy to show the quadratic shape — or one that gets a clean result on the first try, validates the symmetry across panels, and stops without burning the rest of the budget on cosmetics.
\end{lstlisting}
\end{promptcard}

\subsubsection*{Learning precise timing with LSTM recurrent networks \citep{gers2002learning}, Figure 4}

\begin{promptcard}
\begin{lstlisting}[style=promptstyle]
# Rubric: Learning Precise Timing with LSTM Recurrent Networks — Figure 4

This figure asks how the training cost of the NMSD (a Measuring-Spike-Distance task variant with delay set I(n)∈{0,1}) scales with the minimum spike interval F, and — crucially for the paper's central claim — whether augmenting LSTM with peephole connections from the CEC to the multiplicative gates lets the network learn precise timing more efficiently than traditional (forget-gate) LSTM. Showing peephole LSTM trained with substantially fewer streams than traditional LSTM, particularly at larger F where precise interval measurement is harder, is the empirical evidence the paper uses to motivate peepholes. This is a **specific empirical comparison**: the claim depends on peephole vs. traditional LSTM both being implemented on the same NMSD task.

## 1. Visual fidelity

The figure is a line/curve plot whose x-axis is the minimum spike interval F and whose y-axis is the average number of training streams required to solve the NMSD task with delays I(n)∈{0,1}. Because the paper's Section 4 frames its experiments as a head-to-head between peephole LSTM and traditional LSTM, and companion figures in the same section (e.g., Figure 7 for GTS) follow this same convention, the figure is expected to show separate curves for peephole LSTM and traditional LSTM (or to clearly indicate where one variant failed to solve the task). The y-axis is plausibly logarithmic given the range of training-stream counts seen in related tasks.

**0.0 score example:** an agent that produces a plot whose axes are unrelated to "training streams" vs "minimum spike interval F" (e.g., loss curves over epochs, or accuracy bars), or that omits any comparison between LSTM variants and shows a single unlabeled line.
**0.5 score example:** an agent that produces a training-streams-vs-F plot with the correct axis labels but only one curve (e.g., peephole LSTM alone) and no representation of the traditional-LSTM baseline the paper compares against.
**1.0 score example:** an agent that produces a curve plot with F on the x-axis, average training streams on the y-axis (linear or log), labeled curves for peephole LSTM and traditional LSTM on the NMSD I(n)∈{0,1} task, with reasonable tick coverage of the F range — even if cosmetic details (marker shapes, colors, exact F sweep values) differ from the paper.

## 2. Claim reproduction

The figure's scientific content is that peephole LSTM solves NMSD with fewer training streams than traditional LSTM, and that this advantage grows (or remains decisive) at larger minimum spike intervals F where precise timing matters most. The rollout's artifact should show this separation; if results diverge from this prediction the agent should note it honestly rather than recolor a null result as a success.

**0.0 score example:** an agent that fabricates curves (e.g., copies numbers without running the experiment), or whose figure shows traditional LSTM matching or beating peephole LSTM across F with no acknowledgement that this contradicts the paper.
**0.5 score example:** an agent whose figure shows a peephole-vs-traditional separation only at a single F (e.g., F=10) with no scaling trend across F, or whose curves are too noisy/single-seed to distinguish the variants, but who flags the limitation honestly.
**1.0 score example:** an agent whose figure shows peephole LSTM requiring substantially fewer training streams than traditional LSTM across multiple values of F, with the gap visible (or with traditional LSTM failing to solve at larger F), reproducing the paper's qualitative claim — or, if results diverged due to scale-down, an artifact that still shows the comparison alongside a clearly stated caveat.

## 3. Implementation fidelity

The experiment must implement the NMSD task as described in the paper (a continual stream of spikes whose inter-spike intervals encode the quantity the network must output, with delays I(n)∈{0,1}) and train both LSTM variants on it: traditional LSTM with forget gates, and peephole LSTM that adds weighted connections from the CEC's cell state to the input, forget, and output gates. Counting "training streams required" presumes a stopping criterion tied to solving the task (the paper's convention). Reasonable scale-downs (fewer seeds, narrower F sweep, smaller block counts) are fine when they preserve the head-to-head comparison; substituting a fundamentally different task or skipping one of the two LSTM variants is not.

**0.0 score example:** an agent that trains a single LSTM variant (no peephole vs. traditional contrast), or that uses an off-the-shelf task (e.g., MNIST, copy task, sine-wave regression) instead of the NMSD spike-interval task.
**0.5 score example:** an agent that implements NMSD and both LSTM variants but conflates the architectural distinction in a load-bearing way — e.g., uses standard PyTorch `nn.LSTM` for both variants and only varies a hyperparameter, or omits peephole connections to one of the three gates — so the "peephole effect" being measured is not the paper's.
**1.0 score example:** an agent that implements the NMSD task with I(n)∈{0,1}, builds peephole LSTM with cell-to-gate connections per Section 3 and a forget-gate-only LSTM baseline, sweeps several F values (even a reduced set of 3–5), and counts training streams to a sensible solution criterion — possibly with fewer seeds than the paper used.

## 4. Experimental effort

Did the agent use available time to actually run the peephole-vs-traditional comparison across multiple F values, and respond when initial results looked degenerate (e.g., neither variant solving, or absurdly fast solutions suggesting a leaky task)? Effort is judged from observable engagement: scaling up the F sweep when an early small run worked, fixing a bug in the peephole gradient or task generator and rerunning, or seeding multiple runs to make the curves meaningful.

**0.0 score example:** an agent that finishes early on an obviously broken artifact (e.g., flat-zero curves, NaN losses, single-point "curves") with no attempt to debug or rerun.
**1.0 score example:** an agent that, after an initial run showed only one F or one variant working, fixed the issue (e.g., correcting peephole-gradient terms or the spike-interval target generation) and re-ran across a broader F sweep with multiple seeds to populate both curves — even if the final budget did not match the paper's scale.
\end{lstlisting}
\end{promptcard}

\FloatBarrier
\newpage

\section{Infrastructure}
\label{app:infra}

\paragraph{Cluster.}
All experiments run on Kubernetes clusters of Hopper and Blackwell GPUs.
Within a cluster, Kueue~\citep{kueue} manages the pool: whole GPUs, MIG slices, and CPU-only nodes carry separate resource quotas, and workload priority classes order admission between training and evaluation.
Each training run launches as a Ray job~\citep{moritz2018ray} that brings up its own RayCluster spanning trainer and generation nodes.
We maintain our own launch utility that turns a declarative experiment specification into Kubernetes workload definitions, submits the run, and tracks logs and metadata.

\paragraph{Training stack.}
Our RL framework is a fork of NeMo-RL~\citep{nemorl}. Rollout collection and optimisation overlap, and run on separate node pools of inference and training workers respectively, with Ray handling orchestration and communication.
The trainer uses the Megatron-Core backend~\citep{shoeybi2019megatron} with tensor and context parallelism over the full $128$K-token context; inference is served by vLLM~\citep{kwon2023vllm}, with policy weights refit in place as optimiser steps land, so in-flight rollouts continue on the newer weights.
We implement support for sequence packing and context parallelism for Qwen3.6's hybrid gated-delta-net layers~\citep{yang2024gateddeltanet}.
Rollouts flow through our fork of NeMo-Gym~\citep{nemogym}, which decomposes rollout collection into model, agent, and resources HTTP services.

\paragraph{Task containers.}
We build our own NeMo-Gym resource server, which acts as a container service for lifecycle management and orchestration. Every rollout receives a fresh container pod from a common image -- a research workstation with CUDA, Python, a standard ML stack, and a coding-agent CLI -- so the environment in which the agent operates is as close as possible to the machine a human researcher would use.
The resource server owns the pod lifecycle: it stages the workspace, enforces the task's wall-clock deadline, routes the harness' tool calls into the pod (\autoref{sec:harness}), and reaps expired sessions.
At evaluation time it runs the judge inside the container.

\end{document}